\documentclass[10pt,twocolumn,letterpaper]{article}

\usepackage{cvpr}
\usepackage{times}
\usepackage{epsfig}
\usepackage{graphicx}
\usepackage{float}
\usepackage{amsmath}
\usepackage{amssymb}

\usepackage[pagebackref=false,breaklinks=true,colorlinks,urlcolor=blue,citecolor=blue,linkcolor=blue,bookmarks=false]{hyperref}

\usepackage{helvet}   %Required
\usepackage{courier}  %Required
\usepackage{xcolor}
\usepackage{booktabs}
\usepackage{multirow}
\usepackage{subcaption}

\DeclareMathOperator*{\argmin}{arg\,min}

\DeclareMathAlphabet      {\mathbfit}{OML}{cmm}{b}{it}

\newcommand{\ra}[1]{\renewcommand{\arraystretch}{#1}}

\renewcommand{\etal}{\textit{et~al}\mbox{.}}
\renewcommand{\eg}{e.g.,\ }
\renewcommand{\ie}{i.e.,\ }

\newcommand{\figref}[1]{Figure~\ref{fig:#1}}

\newlength{\fourimg}

\newlength\paramargin
\newlength\figmargin
\newlength\tablemargin
\newlength\secmargin
\newlength\figcapmargin
\newlength\tablecapmargin

\newcommand{\tb}[1]{\textbf{#1}}

\newcommand{\mpage}[2]
{
\begin{minipage}[b]{#1\linewidth}\centering
#2
\end{minipage}
}

\cvprfinalcopy % *** Uncomment this line for the final submission

\def\cvprPaperID{281} % *** Enter the CVPR Paper ID here

\ifcvprfinal\fi

\begin{document}
\pagenumbering{gobble}
%%%%%%%%% TITLE
\title{CrDoCo: Pixel-level Domain Transfer with Cross-Domain Consistency}

\author{
Yun-Chun~Chen$^{1,2}$ \hspace{20pt} 
Yen-Yu~Lin$^{1}$ \hspace{20pt} 
Ming-Hsuan~Yang$^{3,4}$ \hspace{20pt} 
Jia-Bin~Huang$^{5}$\\
\vspace{2mm} 
$^{1}$Academia Sinica \hspace{12pt} 
$^{2}$National Taiwan University \hspace{12pt} 
$^{3}$UC Merced \hspace{12pt} 
$^{4}$Google \hspace{12pt} 
$^{5}$Virginia Tech
}

\twocolumn[{
\renewcommand\twocolumn[1][]{#1}
\maketitle
\begin{center}
\mpage{0.01}{\raisebox{12pt}{\rotatebox{90}{Semantic seg.}}} 
\mpage{0.315}{\includegraphics[width=\linewidth]{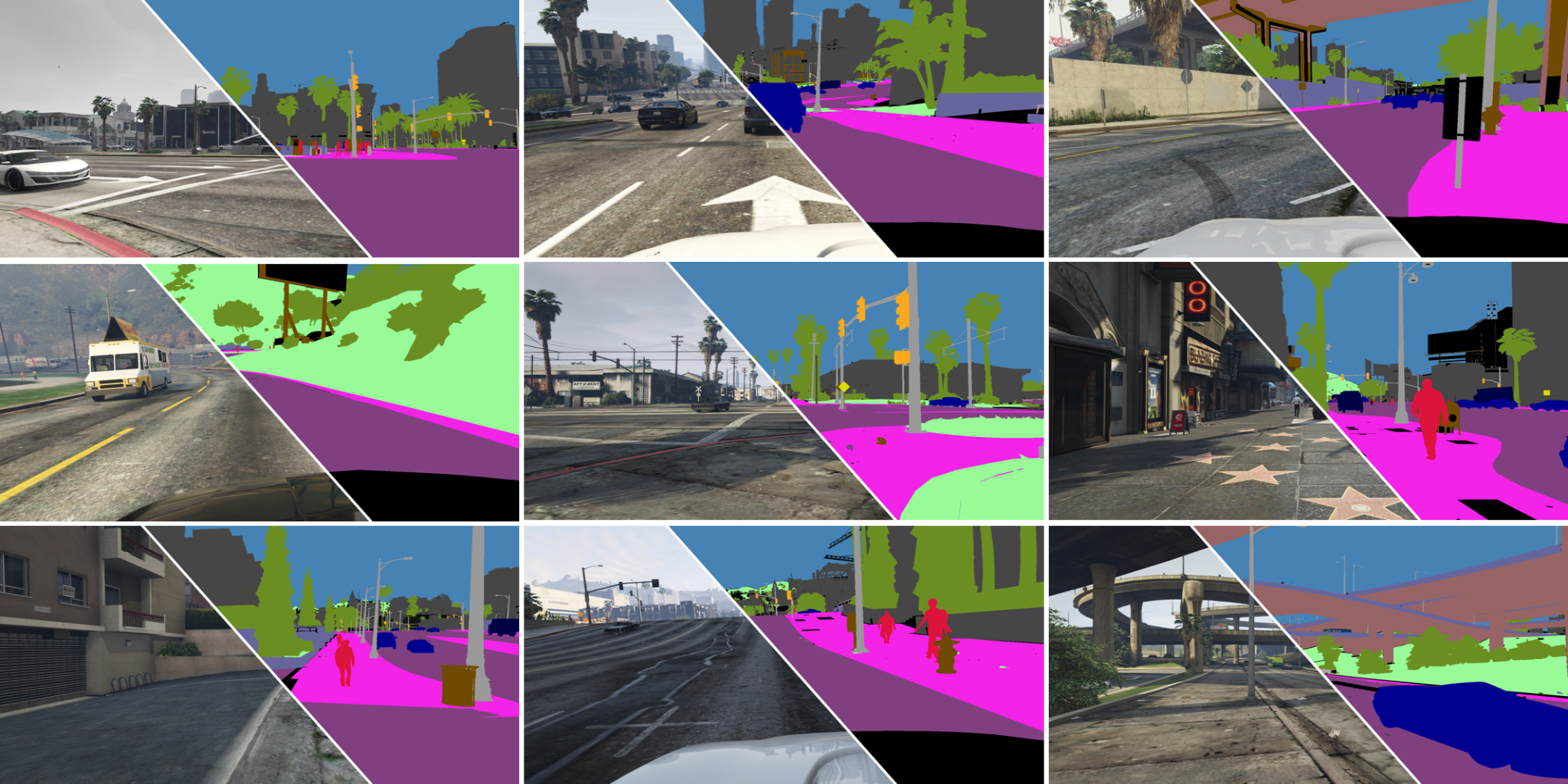}} \hfill
\mpage{0.315}{\includegraphics[width=\linewidth]{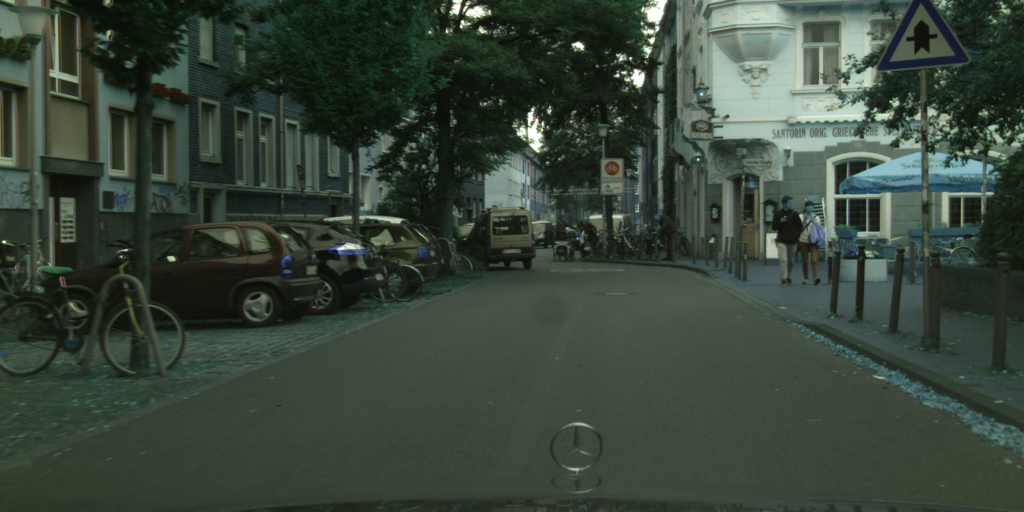}} \hfill
\mpage{0.315}{\includegraphics[width=\linewidth]{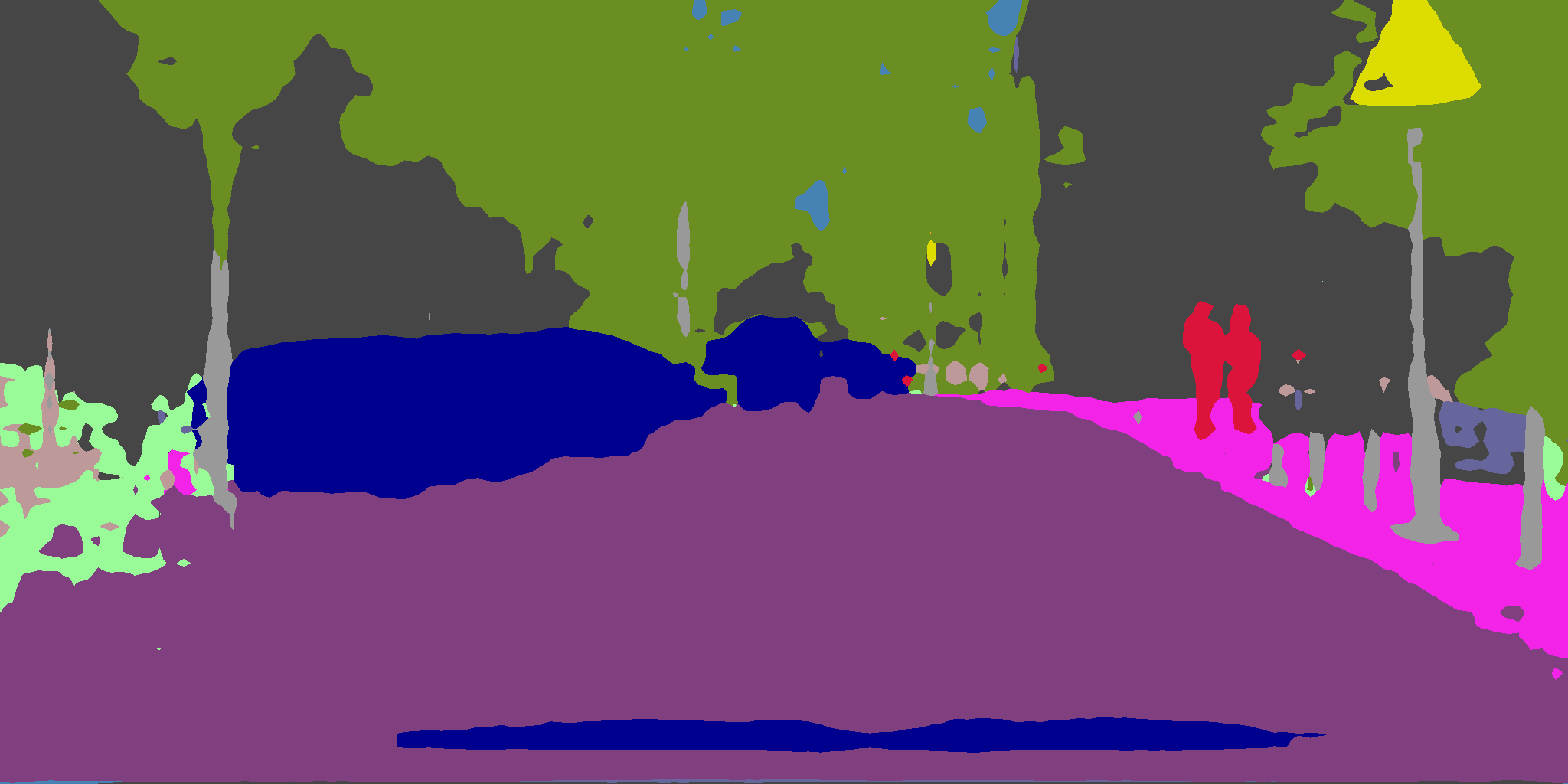}} \\
\vspace{1.0mm}
\mpage{0.01}{\raisebox{10pt}{\rotatebox{90}{Depth prediction}}}
\mpage{0.315}{\includegraphics[width=\linewidth]{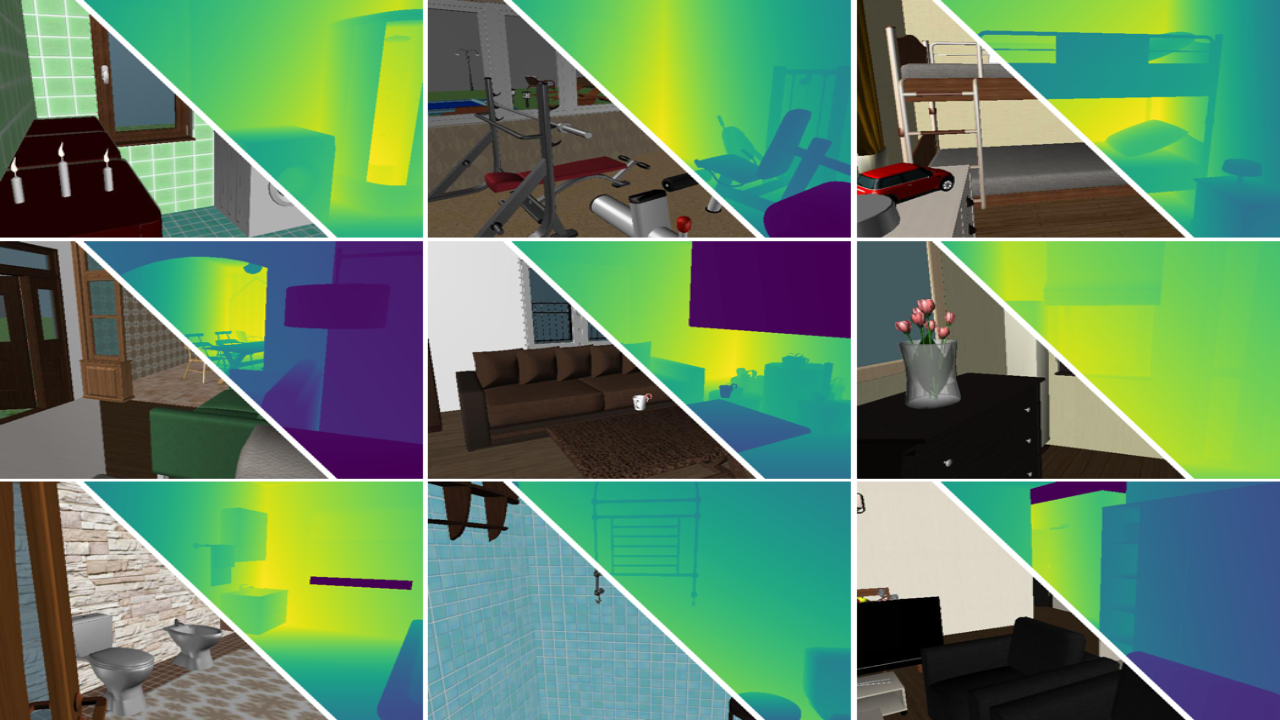}} \hfill
\mpage{0.315}{\includegraphics[width=\linewidth]{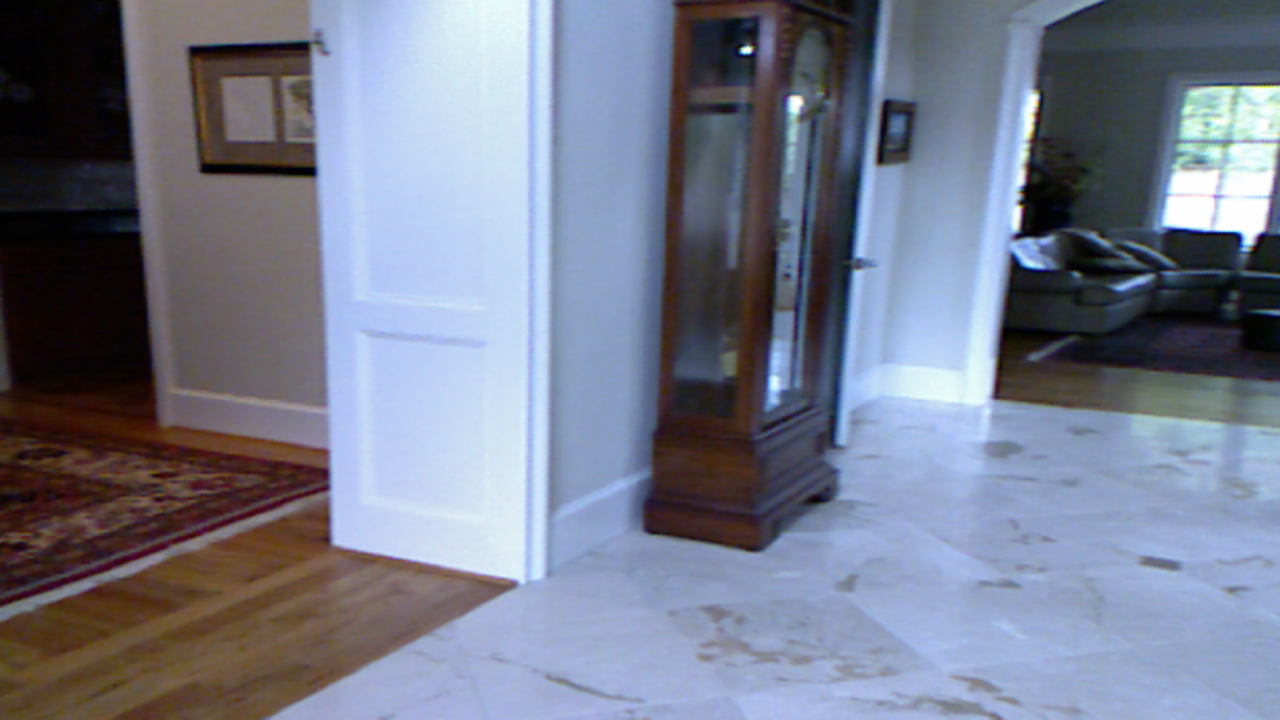}} \hfill
\mpage{0.315}{\includegraphics[width=\linewidth]{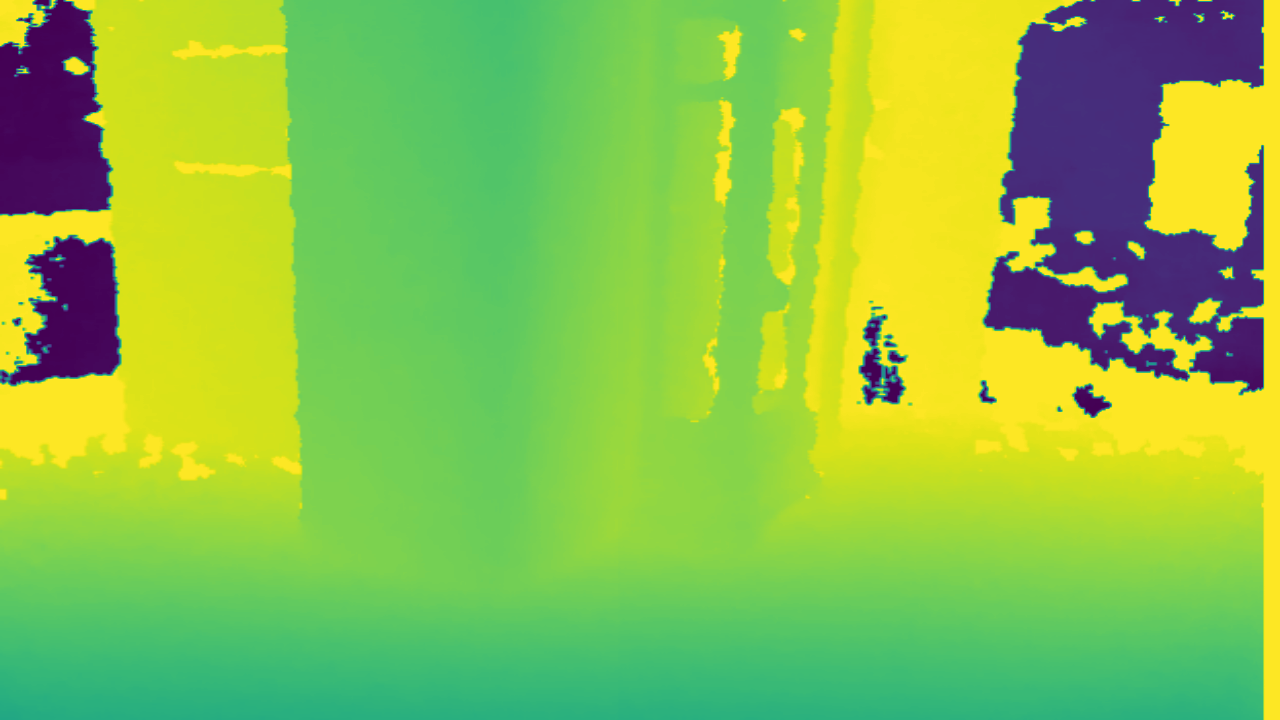}} \\
\vspace{1.0mm}
\mpage{0.01}{\raisebox{0pt}{\rotatebox{90}{Optical flow}}}
\mpage{0.315}{\includegraphics[width=\linewidth]{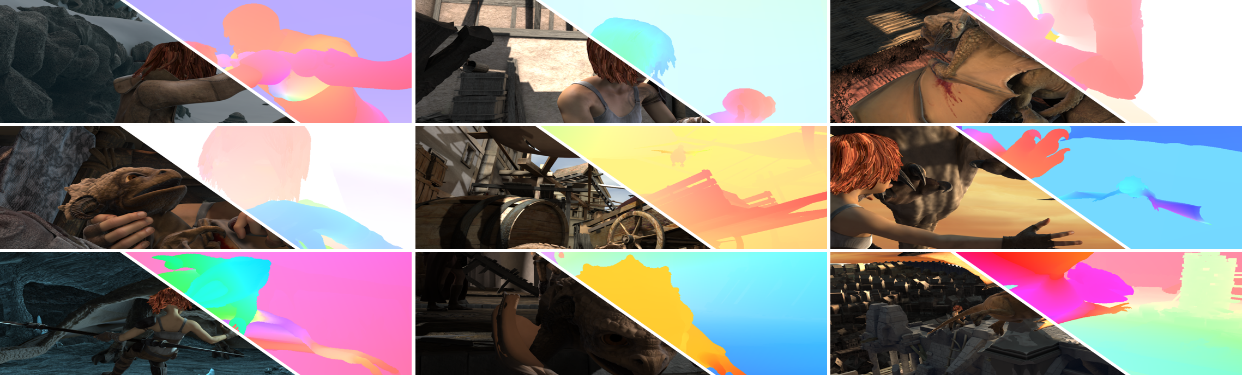}} \hfill
\mpage{0.315}{\includegraphics[width=\linewidth]{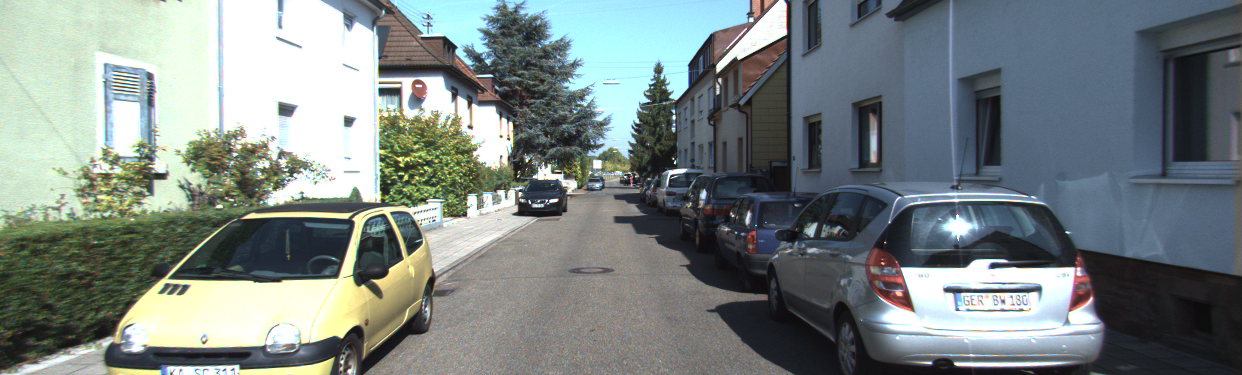}} \hfill
\mpage{0.315}{\includegraphics[width=\linewidth]{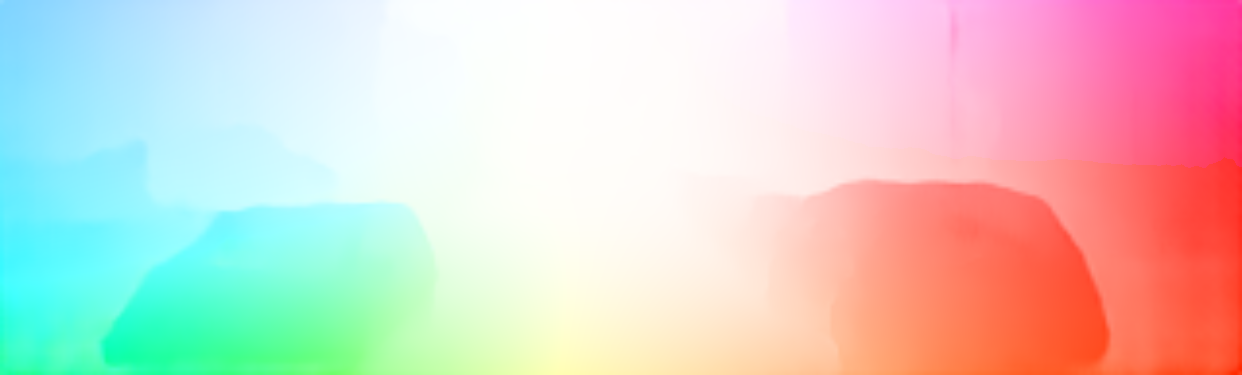}} \\
\mpage{0.01}{}
\mpage{0.315}{Labeled examples (source domain)} \hfill
\mpage{0.315}{Input (target domain)} \hfill
\mpage{0.315}{Output} \\

  \vspace{-2.0mm}
  \captionof{figure}{\textbf{Applications of the proposed method.} Our method has the applications ranging from semantic segmentation (top row), depth prediction (middle row), to optical flow estimation (bottom row).
  %
%  \jb{Getting better! 
%  1) Please see my typesetted example code above. Typset row by row. Not column by column.
%  2) First column. Show a GRID of examples (e.g., 4x6). This is the training data, so we must show multiple of them. For each of them, use the style like \url{https://filebox.ece.vt.edu/~jbhuang/images/projects/SigAsia 2016 Video Completion.gif}. 
%  3) I notice that the size for the input and output have been changed. This creates some distortion (particularly for the 2nd column.)
%  4) Pick examples where our method works well. for example, the semantic segmentation, we probably want to see more variety of labels (e.g., people). For flow, it's better to show large foreground moving object. For depth, it's better to show results without holes. Flow depth, visualize them using color map (instead of grayscale). We typically use ``viridis".
%  }
% \jb{The teaser should give an overall idea of what we are doing. Currently, it's just three applications. I suggest to have the following.
% For each row: 
% - column one: a grid of source image/annotations (implying that we have labeled data in the source)
% - column two: an input image in the target domain (the first row in your current figure).
% - column three: results (the second row).
% %
% Move semantic segmentation, depth, flow to vertical text on the LEFT of the images. 
% At the bottom of each column, subcaption: labeled source dataset, input from target domain, output. 
% % 
% What do you think?
% }
  }
  \label{fig:teaser}
\end{center}
}]

\begin{abstract}
Unsupervised domain adaptation algorithms aim to transfer the knowledge learned from one domain to another (\eg synthetic to real images).
The adapted representations often do not capture pixel-level domain shifts that are crucial for dense prediction tasks (\eg semantic segmentation).
In this paper, we present a novel pixel-wise adversarial domain adaptation algorithm. 
By leveraging image-to-image translation methods for data augmentation, our key insight is that while the translated images between domains may differ in styles, their predictions for the task should be consistent.
We exploit this property and introduce a cross-domain consistency loss that enforces our adapted model to produce consistent predictions.
Through extensive experimental results, we show that our method compares favorably against the state-of-the-art on a wide variety of unsupervised domain adaptation tasks.
\end{abstract}

\section{Introduction} \label{sec:intro}

Deep convolutional neural networks (CNNs) are extremely data hungry. 
However, for many dense prediction tasks (\eg semantic segmentation, optical flow estimation, and depth prediction), collecting large-scale and diverse datasets with pixel-level annotations is difficult since the labeling process is often expensive and labor intensive (see Figure~\ref{fig:teaser}).
Developing algorithms that can transfer the knowledge learned from one labeled dataset (\ie source domain) to another unlabeled dataset (\ie target domain) thus becomes increasingly important.
Nevertheless, due to the domain-shift problem (\ie the domain gap between the source and target datasets), the learned models often fail to generalize well to new datasets.

To address these issues, several unsupervised domain adaptation methods have been proposed to align data distributions between the source and target domains.
Existing methods either apply feature-level~\cite{sun2016deep,long2015learning,tzeng2017adversarial,tsai2018learning,hoffman2016fcns,hoffman2017cycada} or pixel-level~\cite{bousmalis2017unsupervised,shrivastava2017learning,dundar2018domain,hoffman2017cycada} adaptation techniques to minimize the domain gap between the source and target datasets.
However, aligning marginal distributions does not necessarily lead to satisfactory performance as there is no explicit constraint imposed on the predictions in the target domain (as no labeled training examples are available).
While several methods have been proposed to alleviate this issue via curriculum learning~\cite{sakaridis2018model,dai2018dark} or self-paced learning~\cite{zou2018domain}, the problem remains challenging since these methods may only learn from cases where the current models perform well.

\begin{figure}[t]
  \centering
  \includegraphics[width=\linewidth]{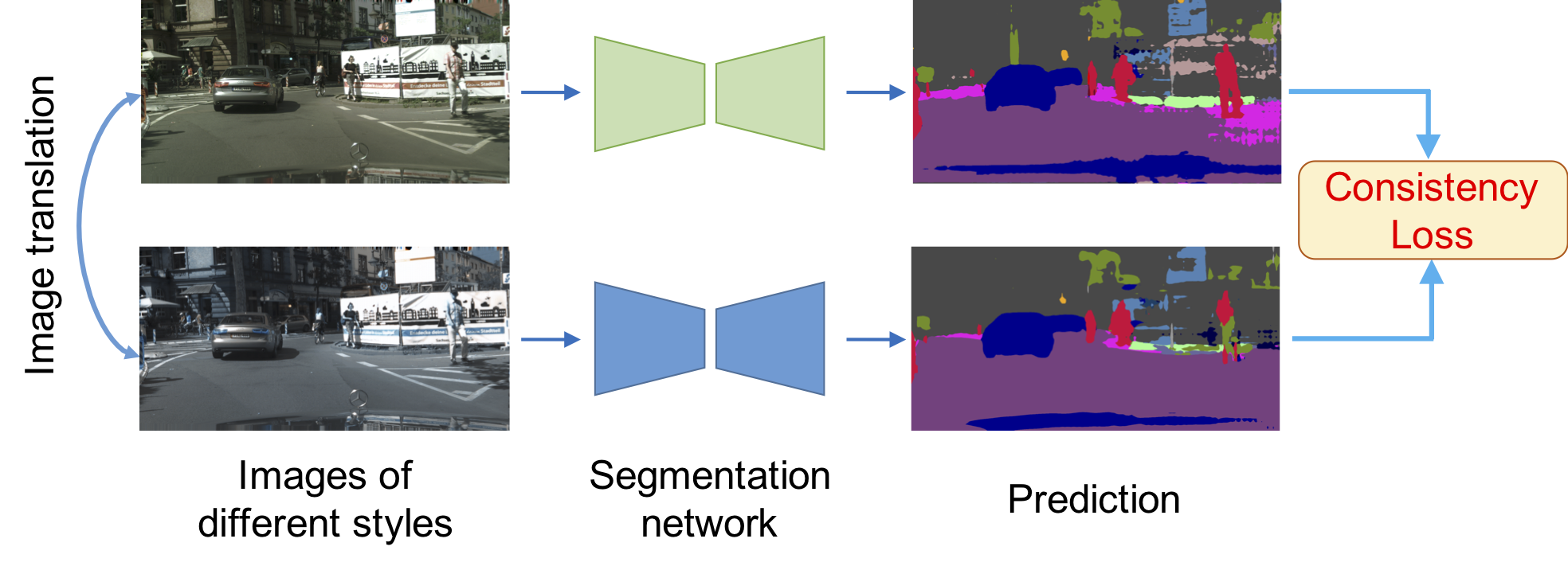}
  \vspace{\figcapmargin}
  \caption{\textbf{Main idea.}
  While images may have different appearances/styles in different domains, their task predictions (\eg semantic segmentation as shown in this example) should be exactly the same.
  Our core idea in this paper is to impose a cross-domain consistency loss between the two task predictions.
  }
  \label{fig:motivation}
  \vspace{\figmargin}
\end{figure}

\vspace{\paramargin}
\paragraph{Our work.} In this paper, we present CrDoCo, a pixel-level adversarial domain adaptation algorithm for dense prediction tasks.
Our model consists of two main modules: 1) an image-to-image translation network and 2) two domain-specific task networks (one for source and the other for target). 
The image translation network learns to translate images from one domain to another such that the translated images have a similar distribution to those in the translated domain.
The domain-specific task network takes images of source/target domain as inputs to perform dense prediction tasks.
As illustrated in \figref{motivation}, our core idea is that while the original and the translated images in two different domains may have different styles, their predictions from the respective domain-specific task network should be exactly the same. 
We enforce this constraint using a \emph{cross-domain consistency loss} that provides additional supervisory signals for facilitating the network training, allowing our model to produce consistent predictions. 
We show the applicability of our approach to multiple different tasks in the unsupervised domain adaptation setting.

\vspace{\paramargin}
\paragraph{Our contributions.}
First, we present an adversarial learning approach for unsupervised domain adaptation which is applicable to a wide range of dense prediction tasks.
Second, we propose a cross-domain consistency loss that provides additional supervisory signals for network training, resulting in more accurate and consistent task predictions.
Third, extensive experimental results demonstrate that our method achieves the state-of-the-art performance against existing unsupervised domain adaptation techniques.
Our source code is available at 
\href{https://yunchunchen.github.io/CrDoCo/}{https://yunchunchen.github.io/CrDoCo/}

\section{Related Work} \label{sec:related}

\paragraph{Unsupervised domain adaptation.} Unsupervised domain adaptation methods can be categorized into two groups: 1) feature-level adaptation and 2) pixel-level adaptation.
Feature-level adaptation methods aim at aligning the feature distributions between the source and target domains through measuring the correlation distance~\cite{sun2016deep}, minimizing the maximum mean discrepancy~\cite{long2015learning}, or applying adversarial learning strategies~\cite{tzeng2017adversarial,tsai2018learning} in the feature space.
In the context of image classification, several methods~\cite{ganin2014unsupervised,ganin2016domain,long2015learning,long2016unsupervised,tzeng2015simultaneous,tzeng2017adversarial,chen2019learning} have been developed to address the domain-shift issue.
For semantic segmentation tasks, existing methods often align the distributions of the feature activations at multiple levels~\cite{hoffman2016fcns,Huang_2018_ECCV,tsai2018learning}.
Recent advances include applying class-wise adversarial learning~\cite{chen2017no} or leveraging self-paced learning policy~\cite{zou2018domain} for adapting synthetic-to-real or cross-city adaptation~\cite{chen2017no}, adopting curriculum learning for synthetic-to-real foggy scene adaptation~\cite{sakaridis2018model}, or progressively adapting models from daytime scene to nighttime~\cite{dai2018dark}.
Another line of research focuses on pixel-level adaptation~\cite{bousmalis2017unsupervised,shrivastava2017learning,dundar2018domain}.
These methods address the domain gap problem by performing data augmentation in the target domain via image-to-image translation~\cite{bousmalis2017unsupervised,shrivastava2017learning} or style transfer~\cite{dundar2018domain} methods.

Most recently, a number of methods tackle joint feature-level and pixel-level adaptation in image classification~\cite{hoffman2017cycada,li2019recover}, semantic segmentation~\cite{hoffman2017cycada}, and single-view depth prediction~\cite{zheng2018t2net} tasks.
These methods~\cite{hoffman2017cycada,zheng2018t2net} utilize image-to-image translation networks (\eg the CycleGAN~\cite{zhu2017unpaired}) to translate images from source domain to target domain with pixel-level adaptation.
The translated images are then passed to the task network followed by a feature-level alignment.

While both feature-level and pixel-level adaptation have been explored, aligning the marginal distributions without enforcing explicit constraints on target predictions would not necessarily lead to satisfactory performance.
Our model builds upon existing techniques for feature-level and pixel-level adaptation~\cite{hoffman2017cycada,zheng2018t2net}.
The key difference lies in our cross-domain consistency loss that explicitly penalizes inconsistent predictions by the task networks. 

\vspace{\paramargin}
\paragraph{Cycle consistency.} Cycle consistency constraints have been successfully applied to various problems. In image-to-image translation, enforcing cycle consistency allows the network to learn the mappings without paired data~\cite{zhu2017unpaired,DRIT}. In semantic matching, cycle or transitivity based consistency loss help regularize the network training~\cite{Multi-match,FlowWeb,WeakMatchNet,chen2019show}. In motion analysis, forward-backward consistency check can be used for detecting occlusion~\cite{meister2018unflow,lai2018learning,zou2018df} or learning visual correspondence~\cite{wang2019learning}.
Similar to the above methods, we show that enforcing two domain-specific networks to produce consistent predictions leads to substantially improved performance. 

\vspace{\paramargin}
\paragraph{Learning from synthetic data.}
Training the model on large-scale synthetic datasets has been extensively studied in semantic segmentation~\cite{tobin2017domain,tsai2018learning,hoffman2016fcns,hoffman2017cycada,dundar2018domain,Huang_2018_ECCV,sakaridis2018model,sankaranarayanan2018learning,zou2018domain}, multi-view stereo~\cite{huang2018deepmvs}, depth estimation~\cite{zheng2018t2net}, optical flow~\cite{sun2018pwc,ilg2017flownet,lai2017semi}, amodal segmentation~\cite{Hu2019SAILVOS}, and object detection~\cite{dundar2018domain,Peng2018Syn2RealAN}.
In our work, we show that the proposed cross-domain consistency loss can be applied not only to synthetic-to-real adaptation but to real-to-real adaptation tasks as well.

\begin{figure*}[t]
  \centering
  \includegraphics[width=\linewidth]{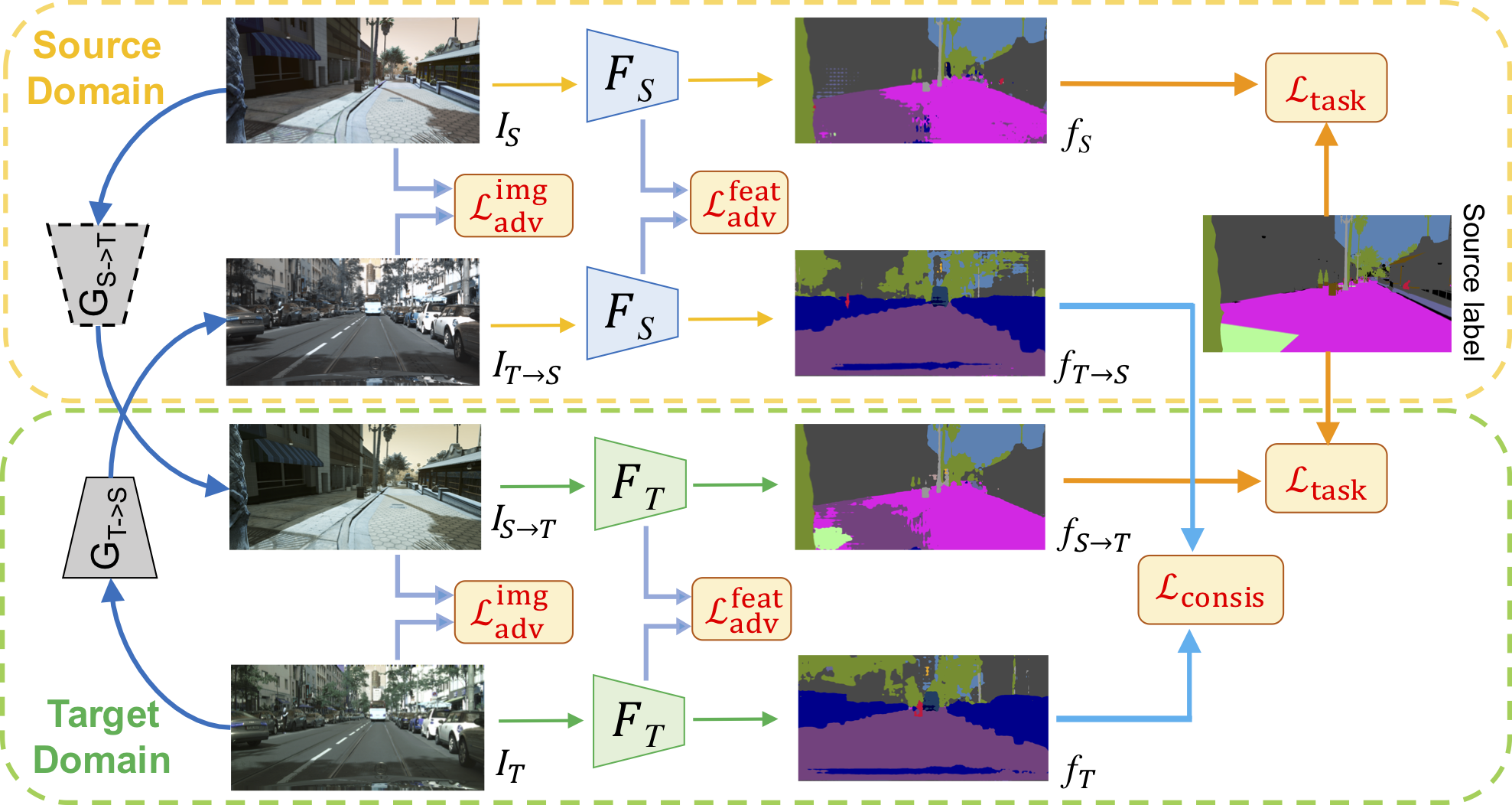}
  \vspace{\figcapmargin}
  \caption{\textbf{Overview of the proposed method.}
  Our model is composed of two main modules: an image translation network (highlighted in gray) and two domain-specific task networks (highlighted in blue and green, respectively).
  The image translation network learns to translate input images from one domain to the other.
  The input and the translated images are then fed to their corresponding domain-specific task networks to perform task predictions.
  Our main contribution lies in the use of cross-domain consistency loss $\mathcal{L}_\textrm{consis}$ for regularizing the network training. 
  }
  \label{fig:model}
  \vspace{\figmargin}
\end{figure*}

\section{Method} \label{sec:method}

In this section, we first provide an overview of our approach.
We then describe the proposed loss function for enforcing cross-domain consistency on dense prediction tasks.
Finally, we describe other losses that are adopted to facilitate network training.

\subsection{Method overview}

We consider the task of unsupervised domain adaptation for dense prediction tasks.
In this setting, we assume that we have access to a source image set $X_S$, a source label set $Y_S$, and an unlabeled target image set $X_T$.
Our goal is to learn a task network $F_T$ that can reliably and accurately predict the dense label for each image in the target domain.

To achieve this task, we present an end-to-end trainable network which is composed of two main modules: 
1) the image translation network $G_{S \rightarrow T}$ and $G_{T \rightarrow S}$ and 
2) two domain-specific task networks $F_S$ and $F_T$.
The image translation network translates images from one domain to the other.
The domain-specific task network takes input images to perform the task of interest.

As shown in Figure~\ref{fig:model}, the proposed network takes an image $I_S$ from the source domain and another image $I_T$ from the target domain as inputs.
We first use the image translation network to obtain the corresponding translated images $I_{S \rightarrow T} = G_{S \rightarrow T} (I_S)$ (in the target domain) and $I_{T \rightarrow S} = G_{T \rightarrow S} (I_T)$ (in the source domain).
We then pass $I_S$ and $I_{T \rightarrow S}$ to $F_S$, $I_T$ and $I_{S \rightarrow T}$ to $F_T$ to obtain their task predictions.

\subsection{Objective function}
The overall training objective $\mathcal{L}$ for training the proposed network consists of five loss terms.
First, the image-level adversarial loss $\mathcal{L}_\mathrm{adv}^\mathrm{img}$ aligns the image distributions between the translated images and the images in the corresponding domain.
Second, the reconstruction loss $\mathcal{L}_\mathrm{rec}$ regularizes the image translation network $G_{S \rightarrow T}$ and $G_{T \rightarrow S}$ to perform self-reconstruction when translating an image from one domain to another followed by a reverse translation.
Third, the feature-level adversarial loss $\mathcal{L}_\mathrm{adv}^\mathrm{feat}$ aligns the distributions between the feature representations of the translated images and the images in the same domain.
Fourth, the task loss $\mathcal{L}_\mathrm{task}$ guides the two domain-specific task networks $F_S$ and $F_T$ to perform dense prediction tasks.
Fifth, the cross-domain consistency loss $\mathcal{L}_\mathrm{consis}$ enforces consistency constraints on the task predictions.
Such a cross-domain loss couples the two domain-specific task networks $F_S$ and $F_T$ during training and provides supervisory signals for the unlabeled target domain image $I_T$ and its translated one $I_{T \rightarrow S}$.
Specifically, the training objective $\mathcal{L}$ is defined as
\vspace{-2mm}
\begin{equation}
  \begin{split}
  \mathcal{L} & = \mathcal{L}_\mathrm{task} + \lambda_\mathrm{consis} \cdot \mathcal{L}_\mathrm{consis} + \lambda_\mathrm{rec} \cdot \mathcal{L}_\mathrm{rec} \\
  & + \lambda_\mathrm{img} \cdot \mathcal{L}_\mathrm{adv}^\mathrm{img} + \lambda_\mathrm{feat} \cdot \mathcal{L}_\mathrm{adv}^\mathrm{feat},
  \end{split}
  \label{eq:obj}
  \vspace{-2.0mm}
\end{equation}
where $\lambda_\mathrm{consis}$, $\lambda_\mathrm{rec}$, $\lambda_\mathrm{img}$, and $\lambda_\mathrm{feat}$ are the hyper-parameters used to control the relative importance of the respective loss terms.
Below we outline the details of each loss function.

\subsection{Cross-domain consistency loss $\mathcal{L}_\mathrm{consis}$}

Since we do not have labeled data in the target domain, to allow our model to produce accurate task predictions on unlabeled data, we first generate a translated version of $I_T$ (\ie $I_{T \rightarrow S}$) by passing $I_T$ to the image translation network $G_{T \rightarrow S}$ (\ie $I_{T \rightarrow S} = G_{T \rightarrow S} (I_T)$).
Our key insight is that while $I_T$ (belongs to the target domain) and $I_{T \rightarrow S}$ (belongs to the source domain) may differ in appearance or styles, these two images should have the same task prediction results (\ie $F_T(I_T)$ and $F_S(I_{T \rightarrow S})$ should be exactly the same).
We thus propose a cross-domain consistency loss $\mathcal{L}_\mathrm{consis}$ that bridges the outputs of the two domain-specific task networks (\ie $F_S$ and $F_T$).
The loss enforces the consistency between the two task predictions $F_T(I_T)$ and $F_S(I_{T \rightarrow S})$.
For semantic segmentation task, we compute the bi-directional KL divergence loss and define the cross-domain consistency loss for semantic segmentation $\mathcal{L}_\mathrm{consis}$ task as
\begin{equation}
  \begin{split}
  &\mathcal{L}_\mathrm{consis}(X_T; G_{S \rightarrow T}, G_{T \rightarrow S}, F_S, F_T) \\
  = & - \mathbb{E}_{I_T \sim X_T} \sum_{h,w,c}^{}{f_{T \rightarrow S}(h,w,c) \log \bigg( f_T(h,w,c) \bigg)} \\
  & - \mathbb{E}_{I_T \sim X_T} \sum_{h,w,c}^{}{f_T(h,w,c) \log \bigg( f_{T \rightarrow S}(h,w,c) \bigg)}, \\
  \end{split}
  \label{eq:cross_domain_loss_seg}
\end{equation}
where $f_T = F_T(I_T)$ and $f_{T \rightarrow S} = F_S(I_{T \rightarrow S})$ are the task predictions for $I_T$ and $I_{T \rightarrow S}$, respectively, while $c$ denotes the number of classes.

As our task models produce different outputs for different tasks, our cross-domain consistency loss $\mathcal{L}_\mathrm{consis}$ is \emph{task-dependent}.
For depth prediction task, we use the $\ell_1$ loss for the cross-domain consistency loss $\mathcal{L}_\mathrm{consis}$.
For optical flow estimation task, the cross-domain consistency loss $\mathcal{L}_\mathrm{consis}$ computes the endpoint error between the two task predictions.

\subsection{Other losses}

In addition to the proposed cross-domain consistency loss $\mathcal{L}_\mathrm{consis}$, we also adopt several other losses introduced in \cite{hoffman2017cycada,zheng2018t2net,zhu2017unpaired} to facilitate the network training. 

\vspace{\paramargin}
\paragraph{Task loss $\mathcal{L}_\mathrm{task}$.}
To guide the training of the two task networks $F_S$ and $F_T$ using labeled data, for each image-label pair ($I_S$, $y_s$) in the source domain, we first translate the source domain image $I_S$ to $I_{S \rightarrow T}$ by passing $I_S$ to $G_{S \rightarrow T}$ (\ie $I_{S \rightarrow T} = G_{S \rightarrow T} (I_S)$).
Similarly, images before and after translation should have the same ground truth label.
Namely, the label for $I_{S \rightarrow T}$ is identical to that of $I_S$ which is $y_s$.

We can thus define the task loss $\mathcal{L}_\mathrm{task}$ for training the two domain-specific task networks $F_S$ and $F_T$ using \emph{labeled} data.
For semantic segmentation, we calculate the cross-entropy loss between the task predictions and the corresponding ground truth labels as our task loss $\mathcal{L}_\mathrm{task}$.
Likewise, the task loss $\mathcal{L}_\mathrm{task}$ is also task dependent.
We use $\ell_1$ loss for depth prediction task and endpoint error for optical flow estimation.

\vspace{\paramargin}
\paragraph{Feature-level adversarial loss $\mathcal{L}_\mathrm{adv}^\mathrm{feat}$.}
In addition to imposing cross-domain consistency and task losses, we apply two feature-level discriminators $D_S^\mathrm{feat}$ (for source domain) and $D_T^\mathrm{feat}$ (for target domain)~\cite{zhu2017unpaired}.
The discriminator $D_S^\mathrm{feat}$ helps align the distributions between the feature maps of $I_S$ (\ie $f_S$) and $I_{T \rightarrow S}$ (\ie $f_{T \rightarrow S}$).
To achieve this, we define the feature-level adversarial loss in the source domain as
\begin{equation}
  \begin{split}
  \mathcal{L}_\mathrm{adv}^\mathrm{feat}&~(X_S, X_T; G_{T \rightarrow S}, \mathcal{F}_S, D_S^\mathrm{feat}) \\
  = &~ \mathbb{E}_{I_S \sim X_S}[\log(D_S^\mathrm{feat}(f_S))] \\
  + &~ \mathbb{E}_{I_T \sim X_T}[\log(1 - D_S^\mathrm{feat}(f_{T \rightarrow S}))]. \\
  \end{split}
  \label{eq:adv_loss_feature}
\end{equation}

Similarly, $D_T^\mathrm{feat}$ aligns the distributions between $f_T$ and $f_{S \rightarrow T}$.
This corresponds to another feature-level adversarial loss in the target domain as $\mathcal{L}_\mathrm{adv}^\mathrm{feat}(X_T, X_S; G_{S \rightarrow T}, \mathcal{F}_T, D_T^\mathrm{feat})$.

\vspace{\paramargin}
\paragraph{Image-level adversarial loss $\mathcal{L}_\mathrm{adv}^\mathrm{img}$.}
In addition to feature-level adaptation, we also consider image-level adaptation between the translated images and those in the corresponding domain.
Similar to Zhu~\etal~\cite{zhu2017unpaired}, we deploy two image-level discriminators $D_S^\mathrm{img}$ (for source domain) and $D_T^\mathrm{img}$ (for target domain).
The $D_S^\mathrm{img}$ aims at aligning the distributions between the image $I_S$ and the translated one $I_{T \rightarrow S}$.
To accomplish this, we define the image-level adversarial loss in the source domain as
\begin{equation}
  \begin{split}
  \mathcal{L}_\mathrm{adv}^\mathrm{img}&~(X_S, X_T; G_{T \rightarrow S}, D_S^\mathrm{img}) \\
  = &~ \mathbb{E}_{I_S \sim X_S}[\log(D_S^\mathrm{img}(I_S))] \\
  + &~ \mathbb{E}_{I_T \sim X_T}[\log(1 - D_S^\mathrm{img}(I_{T \rightarrow S}))]. \\
  \end{split}
  \label{eq:adv_loss_image}
\end{equation}

Similarly, we have another image-level adversarial loss in the target domain as $\mathcal{L}_\mathrm{adv}^\mathrm{img}(X_T, X_S; G_{S \rightarrow T}, D_T^\mathrm{img})$.

\vspace{\paramargin}
\paragraph{Reconstruction loss $\mathcal{L}_\mathrm{rec}$.}

Finally, we use an image reconstruction loss $\mathcal{L}_\mathrm{rec}$ to regularize the training of the image translation network. 
We exploit the property that when translating an image from one domain to another followed by performing a reverse translation, we should obtain the same image.
Namely, $G_{T \rightarrow S}(G_{S \rightarrow T}(I_S)) \approx I_S$ for any $I_S$ in the source domain and $G_{S \rightarrow T}(G_{T \rightarrow S}(I_T)) \approx I_T$ for any $I_T$ in the target domain hold.

More precisely, we define the reconstruction loss $\mathcal{L}_\mathrm{rec}$ as
\begin{equation}
  \begin{split}
  \mathcal{L}_\mathrm{rec}&~(X_S, X_T; G_{S \rightarrow T}, G_{T \rightarrow S}) \\
  = &~ \mathbb{E}_{I_S \sim X_S}[\|G_{T \rightarrow S}(G_{S \rightarrow T}(I_S)) - I_S\|_1] \\
  + &~ \mathbb{E}_{I_T \sim X_T}[\|G_{S \rightarrow T}(G_{T \rightarrow S}(I_T)) - I_T\|_1]. \\
  \end{split}
  \label{eq:rec_loss}
\end{equation}

Following Zhu~\etal~\cite{zhu2017unpaired}, we use the $\ell_1$ norm to define the reconstruction loss $\mathcal{L}_\mathrm{rec}$.

Based on the aforementioned loss functions, we aim to solve for a target domain task network $F_T^*$ by optimizing the following min-max problem:

\vspace{-2.0mm}
\begin{equation}
  F_T^* = \argmin_{F_T} \min_{\substack{F_S, \\ G_{S \rightarrow T} \\ G_{T \rightarrow S}}} \max_{\substack{D_S^\mathrm{img}, D_T^\mathrm{img} \\ D_S^\mathrm{feat}, D_T^\mathrm{feat}}} \mathcal{L}.
  \label{eq:min-max}
\end{equation}

Namely, to train our network using labeled source domain images and unlabeled target domain images, we minimize the cross-domain consistency loss $\mathcal{L}_\mathrm{consis}$, the task loss $\mathcal{L}_\mathrm{task}$, and the reconstruction loss $\mathcal{L}_\mathrm{rec}$.
The image-level adversarial loss $\mathcal{L}_\mathrm{adv}^\mathrm{img}$ and the feature-level adversarial loss $\mathcal{L}_\mathrm{adv}^\mathrm{feat}$ are optimized to align the image and feature distributions \emph{within the same domain}.
The proposed cross-domain consistency loss, in contrast, aligns the task predictions in two \emph{different domains}.

\subsection{Implementation details} 

We implement our model using PyTorch.
We use the CycleGAN~\cite{zhu2017unpaired} as our image-to-image translation network $G_{S \rightarrow T}$ and $G_{T \rightarrow S}$.
The structure of the image-level discriminators $D_S^\mathrm{img}$ and $D_T^\mathrm{img}$ consists of four residual blocks, each of which is composed of a convolutional layer followed by a ReLU activation.
For the feature-level discriminators $D_S^\mathrm{feat}$ and $D_T^\mathrm{feat}$, we use the same architecture as Tsai~\etal~\cite{tsai2018learning}.
The image-to-image translation network $G_{S \rightarrow T}$ and $G_{T \rightarrow S}$, and the discriminators $D_S^\mathrm{img}$, $D_T^\mathrm{img}$, $D_S^\mathrm{feat}$, and $D_T^\mathrm{feat}$ are all randomly initialized.
We have a batch size of $1$, a learning rate of $10^{-3}$ with momentum $0.9$, and set the weight decay as $5 \times 10^{-4}$.
Our hyper-parameters setting: $\lambda_\mathrm{consis} = 10$, $\lambda_\mathrm{rec} = 10$, $\lambda_\mathrm{img} = 0.1$, and $\lambda_\mathrm{feat} = 0.001$.
We train our model on a single NVIDIA GeForce GTX $1080$ GPU with $12$ GB memory.

\section{Experimental Results} \label{sec:results}

% In this section, we start with describing the experimental settings and reporting the results with comparisons to the state-of-the-art methods on three different tasks, including semantic segmentation, flow estimation, and depth prediction.
%
% The source code is available at \href{https://yunchunchen.github.io/CrDoCo/}{https://yunchunchen.github.io/CrDoCo/}

\begin{table*}[t]
  \scriptsize
  \ra{1.2}
  \begin{center}
    \caption{\textbf{Experimental results of synthetic-to-real adaptation for semantic segmentation.}
    %
    % Top: \texttt{GTA5} $\rightarrow$ \texttt{Cityscapes}. 
    % %
    % Bottom: \texttt{SYNTHIA} $\rightarrow$ \texttt{Cityscapes}.
    %
    We denote the top results as \tb{bold} and \underline{underlined}.
    % The bold and underlined numbers indicate top two results, respectively.
    %
    }
    \label{table:exp-seg-syn2real}
    \resizebox{\textwidth}{!} 
    {
    \begin{tabular}{l|c|ccccccccccccccccccc|c|c}
    \toprule
    \multicolumn{23}{c}{
    \tb{\texttt{GTA5} $\rightarrow$ \texttt{Cityscapes}}
    } \\
    \midrule
    Method & Backbone & \rotatebox{90}{Road} & \rotatebox{90}{Sidewalk} & \rotatebox{90}{Building} & \rotatebox{90}{Wall} & \rotatebox{90}{Fence} & \rotatebox{90}{Pole} & \rotatebox{90}{Traffic Light} & \rotatebox{90}{Traffic Sign} & \rotatebox{90}{Vegetation} & \rotatebox{90}{Terrain} & \rotatebox{90}{Sky} & \rotatebox{90}{Person} & \rotatebox{90}{Rider} & \rotatebox{90}{Car} & \rotatebox{90}{Truck} & \rotatebox{90}{Bus} & \rotatebox{90}{Train} & \rotatebox{90}{Motorbike} & \rotatebox{90}{Bicycle} & \rotatebox{90}{mean IoU} & \rotatebox{90}{Pixel acc.} \\
    \midrule
    Synth.~\cite{dundar2018domain} & \multirow{10}{*}{DRN-$26$~\cite{yu2017dilated}} & 68.9 & 19.9 & 52.8 & 6.5 & 13.6 & 9.3 & 11.7 & 8.0 & 75.0 & 11.0 & 56.5 & 36.9 & 0.1 & 51.3 & 8.5 & 4.7 & 0.0 & 0.1 & 0.0 & 22.9 & 71.9 \\
    DR~\cite{tobin2017domain} &  & 67.5 & 23.5 & 65.7 & 6.7 & 12.0 & 11.6 & 16.1 & 13.7 & 70.3 & 8.3 & 71.3 & 39.6 & 1.6 & 55.0 & 15.1 & 3.0 & 0.6 & 0.2 & 3.3 & 25.5 & 73.8 \\
    CycleGAN~\cite{zhu2017unpaired} &  & 89.3 & \underline{45.1} & \underline{81.6} & \underline{27.5} & 18.6 & 29.0 & \underline{35.7} & 17.3 & 79.3 & \underline{29.4} & 71.5 & 59.7 & 15.7 & \underline{85.3} & 18.2 & 14.8 & 1.4 & 21.9 & 12.5 & 39.6 & 86.6 \\
    UNIT~\cite{liu2017unsupervised} &  & \underline{90.5} & 38.5 & 81.1 & 23.5 & 16.3 & 30.2 & 25.2 & 18.5 & 79.5 & 26.8 & \underline{77.8} & 59.2 & \underline{17.4} & 84.4 & \underline{22.2} & 16.1 & 1.6 & 16.7 & \underline{16.9} & 39.1 & 87.1 \\
    FCNs ITW~\cite{hoffman2016fcns} &  & 70.4 & 32.4 & 62.1 & 14.9 & 5.4 & 10.9 & 14.2 & 2.7 & 79.2 & 21.3 & 64.6 & 44.1 & 4.2 & 70.4 & 8.0 & 7.3 & 0.0 & 3.5 & 0.0 & 27.1 & - \\
    CyCADA~\cite{hoffman2017cycada} &  & 79.1 & 33.1 & 77.9 & 23.4 & 17.3 & \underline{32.1} & 33.3 & \textbf{31.8} & 81.5 & 26.7 & 69.0 & \underline{62.8} & 14.7 & 74.5 & 20.9 & \underline{25.6} & \textbf{6.9} & 18.8 & \textbf{20.4} & 39.5 & 82.3 \\
    DS~\cite{dundar2018domain} &  & 89.0 & 43.5 & 81.5 & 22.1 & 8.5 & 27.5 & 30.7 & 18.9 & \underline{84.8} & 28.3 & \textbf{84.1} & 55.7 & 5.4 & 83.2 & 20.3 & \textbf{28.3} & 0.1 & 8.7 & 6.2 & 38.3 & \underline{87.2} \\
    GAM~\cite{Huang_2018_ECCV} &  & - & - & - & - & - & - & - & - & - & - & - & - & - & - & - & - & - & - & - & \underline{40.2} & 81.1 \\
    %
    % \midrule
    %
    Ours w/o $\mathcal{L}_\mathrm{consis}$ &  & 89.1 & 44.9 & 80.9 & \underline{27.5} & \underline{18.8} & 30.2 & 35.6 & 17.1 & 79.5 & 27.2 & 71.6 & 59.7 & 16.1 & 84.6 & 18.1 & 14.6 & 1.4 & \underline{22.1} & 10.9 & 39.4 & 85.8 \\
    Ours &  & \textbf{95.1} & \textbf{49.2} & \textbf{86.4} & \textbf{35.2} & \textbf{22.1} & \textbf{36.1} & \textbf{40.9} & \underline{29.1} & \textbf{85.0} & \textbf{33.1} & 75.8 & \textbf{67.3} & \textbf{26.8} & \textbf{88.9} & \textbf{23.4} & 19.3 & \underline{4.3} & \textbf{25.3} & 13.5 & \textbf{45.1} & \textbf{89.2} \\
    \midrule
    Synth.~\cite{zhang2017curriculum} & \multirow{4}{*}{FCN$8$s~\cite{long2015fully}} & 18.1 & 6.8 & 64.1 & 7.3 & 8.7 & \textbf{21.0} & 14.9 & \textbf{16.8} & 45.9 & 2.4 & 64.4 & 41.6 & 17.5 & 55.3 & 8.4 & 5.0 & \textbf{6.9} & 4.3 & 13.8 & 22.3 & - \\
    Curr. DA~\cite{zhang2017curriculum} &  & 74.9 & 22.0 & 71.7 & 6.0 & 11.9 & 8.4 & 16.3 & 11.1 & 75.7 & 13.3 & 66.5 & 38.0 & 9.3 & 55.2 & 18.8 & \underline{18.9} & 0.0 & 16.8 & 16.6 & 28.9 & - \\
    %
%   Synth.~\cite{sankaranarayanan2018learning} & FCN$8$s-VGG$16$~\cite{long2015fully} & 73.5 & 21.3 & 72.3 & 18.9 & 14.3 & 12.5 & 15.1 & 5.3 & \underline{77.2} & 17.4 & 64.3 & 43.7 & 12.8 & 75.4 & \textbf{24.8} & 7.8 & 0.0 & 4.9 & 1.8 & 29.6 & - \\
    %
    LSD~\cite{sankaranarayanan2018learning} &  & \underline{88.0} & \underline{30.5} & \underline{78.6} & \underline{25.2} & \underline{23.5} & 16.7 & \textbf{23.5} & 11.6 & \textbf{78.7} & \underline{27.2} & \underline{71.9} & \underline{51.3} & \underline{19.5} & \textbf{80.4} & \underline{19.8} & 18.3 & 0.9 & \textbf{20.8} & \underline{18.4} & \underline{37.1} & - \\
    %
    % \midrule
    %
    Ours &  & \textbf{89.1} & \textbf{33.2} & \textbf{80.1} & \textbf{26.9} & \textbf{25.0} & \underline{18.3} & \underline{23.4} & \underline{12.8} & \underline{77.0} & \textbf{29.1} & \textbf{72.4} & \textbf{55.1} & \textbf{20.2} & \underline{79.9} & \textbf{22.3} & \textbf{19.5} & \underline{1.0} & \underline{20.1} & \textbf{18.7} & \textbf{38.1} & \textbf{86.3} \\
    \toprule
    \multicolumn{23}{c}{
    \tb{\texttt{SYNTHIA} $\rightarrow$ \texttt{Cityscapes}}} \\
    \midrule
    Synth.~\cite{dundar2018domain} & \multirow{8}{*}{DRN-$26$~\cite{yu2017dilated}} & 28.5 & 10.8 & 49.6 & 0.2 & 0.0 & 18.5 & 0.7 & 5.6 & 65.3 & - & 71.6 & 36.6 & 6.4 & 43.8 & - & 2.7 & - & 0.8 & 10.0 & 18.5 & 54.6 \\
    DR~\cite{tobin2017domain} &  & 31.3 & 16.7 & 59.5 & 2.2 & 0.0 & 19.7 & 0.4 & 6.2 & 64.7 & - & 67.3 & 43.1 & 3.9 & 35.1 & - & 8.3 & - & 0.3 & 5.5 & 19.2 & 57.9 \\
    CycleGAN~\cite{zhu2017unpaired} &  & 58.8 & 20.4 & 71.6 & 1.6 & \underline{0.7} & 27.9 & 2.7 & 8.5 & 73.5 & - & 73.1 & 45.3 & \underline{16.2} & 67.2 & - & \underline{14.9} & - & 7.9 & \underline{24.7} & 27.1 & 71.4 \\
    UNIT~\cite{liu2017unsupervised} &  & 56.3 & 20.6 & \underline{73.2} & 1.8 & 0.3 & 29.0 & \underline{4.0} & \underline{11.8} & 72.2 & - & \underline{74.5} & 50.7 & \textbf{18.4} & 67.3 & - & \textbf{15.1} & - & 6.7 & \textbf{29.5} & 28.0 & 70.8 \\
    FCNs ITW~\cite{hoffman2016fcns} &  & 11.5 & 19.6 & 30.8 & \textbf{4.4} & 0.0 & 20.3 & 0.1 & 11.7 & 42.3 & - & 68.7 & \underline{51.2} & 3.8 & 54.0 & - & 3.2 & - & 0.2 & 0.6 & 17.0 & - \\
    DS~\cite{dundar2018domain} &  & \textbf{67.0} & \textbf{28.0} & \textbf{75.3} & 4.0 & 0.2 & \underline{29.9} & 3.8 & \textbf{15.7} & \textbf{78.6} & - & \textbf{78.0} & \textbf{54.0} & 15.4 & \textbf{69.7} & - & 12.0 & - & \underline{9.9} & 19.2 & 29.5 & \underline{76.5} \\
    %
    % \midrule
    %
    Ours w/o $\mathcal{L}_\mathrm{consis}$ &  & 58.3 & 17.2 & 64.3 & 2.0 & \underline{0.7} & 24.3 & 2.6 & 5.9 & 72.2 & - & 70.8 & 41.9 & 10.3 & 64.2 & - & 12.5 & - & 8.0 & 21.3 & \underline{29.8} & 75.3 \\
    Ours     & & \underline{62.2} & \underline{21.2} & {72.8} & \underline{4.2} & \textbf{0.8} & \textbf{30.1} & \textbf{4.1} & 10.7 & \underline{76.3} & - & {73.6} & {45.6} & 14.9 & \underline{69.2} & - & 14.1 & - & \textbf{12.2} & 23.0 & \textbf{33.4} & \textbf{79.5} \\
    \midrule
    Synth.~\cite{zhang2017curriculum} & \multirow{4}{*}{FCN$8$s~\cite{long2015fully}} & 5.6 & 11.2 & 59.6 & \textbf{8.0} & \textbf{0.5} & 21.5 & 8.0 & 5.3 & 72.4 & - & 75.6 & 35.1 & 9.0 & 23.6 & - & 4.5 & - & 0.5 & \textbf{18.0} & 22.0 & - \\
    Curr. DA~\cite{zhang2017curriculum} &  & 65.2 & 26.1 & 74.9 & 0.1 & \textbf{0.5} & 10.7 & 3.5 & 3.0 & 76.1 & - & 70.6 & 47.1 & 8.2 & 43.2 & - & \underline{20.7} & - & 0.7 & 13.1 & 29.0 & - \\
    %
    % Synth.~\cite{sankaranarayanan2018learning} & FCN$8$s-VGG$16$~\cite{long2015fully} & 30.1 & 17.5 & 70.2 & \underline{5.9} & 0.1 & 16.7 & 9.1 & 12.6 & 74.5 & - & 76.3 & 43.9 & 13.2 & 35.7 & - & 14.3 & - & 3.7 & 5.6 & 26.8 & - \\
    %
    LSD~\cite{sankaranarayanan2018learning} & & \underline{80.1} & \underline{29.1} & \underline{77.5} & 2.8 & \underline{0.4} & \underline{26.8} & \underline{11.1} & \underline{18.0} & \underline{78.1} & - & \underline{76.7} & \underline{48.2} & \underline{15.2} & \textbf{70.5} & - & 17.4 & - & \underline{8.7} & \underline{16.7} & \underline{36.1} & - \\
    %
    % \midrule
    %
    % Ours (init from Synth.~\cite{sankaranarayanan2018learning}) & & \underline{84.2} & \underline{31.1} & \underline{79.2} & 4.2 & \textbf{0.5} & \underline{28.1} & \underline{12.3} & \underline{19.3} & \textbf{80.3} & - & \underline{77.9} & \underline{49.7} & \textbf{16.4} & \underline{70.2} & - & 20.1 & - & \underline{9.9} & \underline{16.8} & \underline{37.5} & - \\
    %
    Ours &  & \textbf{84.9} & \textbf{32.8} & \textbf{80.1} & \underline{4.3} & \underline{0.4} & \textbf{29.4} & \textbf{14.2} & \textbf{21.0} & \textbf{79.2} & - & \textbf{78.3} & \textbf{50.2} & \textbf{15.9} & \underline{69.8} & - & \textbf{23.4} & - & \textbf{11.0} & 15.6 & \textbf{38.2} & \textbf{84.7} \\
    \bottomrule
    \end{tabular}
    }
  \end{center}
  \vspace{2.0\tablemargin}
\end{table*}

\subsection{Semantic segmentation}

We present experimental results for semantic segmentation in two different settings: 
1) \emph{synthetic-to-real}: adapting from synthetic GTA5~\cite{richter2016playing} and SYNTHIA~\cite{ros2016synthia} datasets to real-world images from Cityscapes dataset~\cite{cordts2016cityscapes} and
2) \emph{real-to-real}: adapting the Cityscapes dataset to different cities~\cite{chen2017no}.
%

% First, we show the evaluations of the model trained on synthetic datasets (\eg, GTA5~\cite{richter2016playing} and SYNTHIA~\cite{ros2016synthia}) and test the adapted model on real-world images from the Cityscapes dataset~\cite{cordts2016cityscapes}.
%
% Second, we conduct experiments on the Cross-City dataset~\cite{chen2017no}, where the model is trained on the Cityscapes dataset~\cite{cordts2016cityscapes} and adapt to other cities in the Cross-City dataset without using annotations.

\subsubsection{GTA5 to Cityscapes}

% \vspace{\paramargin}
\paragraph{Dataset.} 
The GTA5 dataset~\cite{richter2016playing} consists of $24,966$ synthetic images with pixel-level annotations of $19$ categories (compatible with the Cityscapes dataset~\cite{cordts2016cityscapes}).
% from the video game based on the city of Los Angeles.
% %
% The ground truth annotations contain $19$ categories and are compatible with the Cityscapes dataset~\cite{cordts2016cityscapes}.
%
Following Hoffman~\etal~\cite{hoffman2017cycada}, we use the GTA5 dataset and adapt the model to the Cityscapes training set with $2,975$ images.

\vspace{\paramargin}
\paragraph{Evaluation protocols.} 
We evaluate our model on the Cityscapes validation set with $500$ images using the mean intersection-over-union (IoU) and the pixel accuracy as the evaluation metrics.

\vspace{\paramargin}
\paragraph{Task network.} 
%
% \revised{
We evaluate our proposed method using two task networks: 1) dilated residual network-$26$ (DRN-$26$)~\cite{yu2017dilated} and 2) FCN$8$s-VGG$16$~\cite{long2015fully}.
For the DRN-$26$, we initialize our task network from Hoffman~\etal~\cite{hoffman2017cycada}.
For the FCN$8$s-VGG$16$, we initialize our task network from Sankaranarayanan~\etal~\cite{sankaranarayanan2018learning}.
% }

\vspace{\paramargin}
\paragraph{Results.}
We compare our approach with the state-of-the-art methods~\cite{tobin2017domain,zhu2017unpaired,liu2017unsupervised,hoffman2016fcns,hoffman2017cycada,dundar2018domain,Huang_2018_ECCV,sankaranarayanan2018learning,zhang2017curriculum}.
The top block of Table~\ref{table:exp-seg-syn2real} presents the experimental results.
%
% \revised{
% When employing the DRN-$26$~\cite{yu2017dilated} to serve as the task network, our results show that we achieve a mean IoU of $45.1$\% and pixel accuracy of $89.2$\%.
% %
% When using the FCN$8$s-VGG$16$~\cite{long2015fully} as the feature backbone, our results show that we achieve a mean IoU of $38.1$\%.
%
Results on both feature backbones show that our method performs favorably against the state-of-the-art methods, outperforming the previous best competitors by $4.9$\% in mean IoU~\cite{Huang_2018_ECCV} when using the DRN-$26$~\cite{yu2017dilated} and $1.0$\% in mean IoU~\cite{sankaranarayanan2018learning} when using FCN$8$s-VGG$16$~\cite{long2015fully}.
We show that the proposed cross-domain consistency loss $\mathcal{L}_\mathrm{consis}$ is critical for the improved performance (\eg adding $\mathcal{L}_\mathrm{consis}$ improves the mean IoU by $5.7$\% and the pixel accuracy by $3.4$\% when adopting the DRN-$26$~\cite{yu2017dilated} as the task network).
% }
%
Figure~\ref{fig:exp-seg} presents an example that demonstrates the effectiveness of the proposed cross-domain consistency loss $\mathcal{L}_\mathrm{consis}$.
We discover that by applying the cross-domain consistency loss $\mathcal{L}_\mathrm{consis}$, our model produces more consistent and accurate results before and after image translation.

\setlength{\fourimg}{0.24\textwidth}
\begin{figure*}[t]
  \centering
  \begin{subfigure}[t]{\fourimg}
    \centering
    \includegraphics[width=\linewidth]{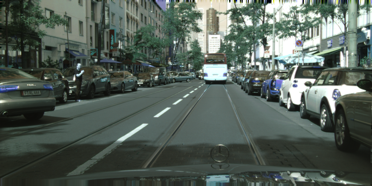}\\
    \vspace{1.0mm}
    \includegraphics[width=\linewidth]{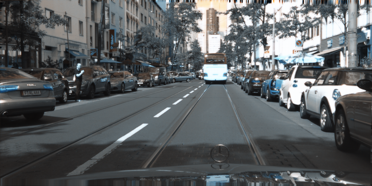}\\
    \caption*{Input images}
  \end{subfigure}
  \hspace{0.01mm}
  \begin{subfigure}[t]{\fourimg}
    \centering
    \includegraphics[width=\linewidth]{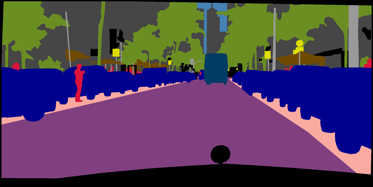}\\
    \vspace{1.0mm}
    \includegraphics[width=\linewidth]{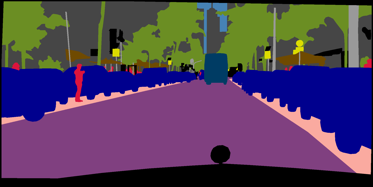}\\
    \caption*{Ground truth}
  \end{subfigure}
  \hspace{0.01mm}
  \begin{subfigure}[t]{\fourimg}
    \centering
    \includegraphics[width=\linewidth]{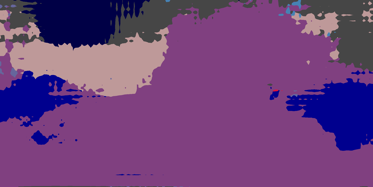}\\
    \vspace{1.0mm}
    \includegraphics[width=\linewidth]{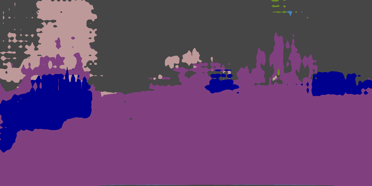}\\
    \caption*{Ours w/o $\mathcal{L}_\mathrm{consis}$}
  \end{subfigure}
  \hspace{0.01mm}
  \begin{subfigure}[t]{\fourimg}
    \centering
    \includegraphics[width=\linewidth]{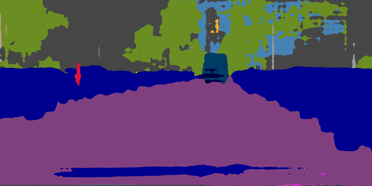}\\
    \vspace{1.0mm}
    \includegraphics[width=\linewidth]{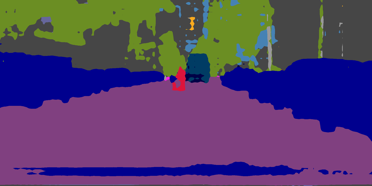}\\
    \caption*{Ours}
  \end{subfigure}
  \vspace{\figcapmargin}
  \caption{\textbf{Visual results of semantic segmentation.} We translate an image from Cityscapes to GTA5. For each input image, we present the segmentation results with and without applying the cross-domain consistency loss.}
  \label{fig:exp-seg}
  \vspace{\figmargin}
\end{figure*}

\begin{table*}[t]
  \scriptsize
  \ra{1.2}
  \begin{center}
    \caption{\textbf{Experimental results of real-to-real adaptation for semantic segmentation. }
    %\jb{Please fix the BOLD and UNDERSCORE}}
    %
    Adaptation: \texttt{Cityscapes} $\rightarrow$ \texttt{Cross-City}.
    %
    %The bold and underlined numbers indicate top two results, respectively.
    %
    }
    \label{table:exp-seg-real2real}
    \vspace{\tablecapmargin}
    \resizebox{\textwidth}{!} 
    {
    \begin{tabular}{l|l|l|ccccccccccccc|c}
    \toprule
    City & Method & Feature backbone & \rotatebox{90}{Road} & \rotatebox{90}{Sidewalk} & \rotatebox{90}{Building} & \rotatebox{90}{Light} & \rotatebox{90}{Sign} & \rotatebox{90}{Vegetation} & \rotatebox{90}{Sky} & \rotatebox{90}{Person} & \rotatebox{90}{Rider} & \rotatebox{90}{Car} & \rotatebox{90}{Bus} & \rotatebox{90}{Motorbike} & \rotatebox{90}{Bicycle} & \rotatebox{90}{mean IoU} \\
    \midrule
    \multirow{6}{*}{Rome} & Cross-City~\cite{chen2017no} & - & 79.5 & 29.3 & 84.5 & 0.0 & 22.2 & 80.6 & 82.8 & 29.5 & 13.0 & 71.7 & 37.5 & 25.9 & 1.0 & 42.9 \\
    %
    % & Synth.~\cite{zou2018domain} & ResNet-38~\cite{he2016deep} & 86.0 & 21.4 & 81.5 & 14.3 & \underline{47.4} & 82.9 & 59.8 & 30.8 & 20.9 & 83.1 & 20.2 & 40.0 & 5.6 & 45.7 \\
    % %
    % & ST~\cite{zou2018domain} & ResNet-38~\cite{he2016deep} & 85.9 & 20.2 & 84.3 & 15.0 & 46.4 & 84.9 & 73.5 & \textbf{48.5} & 21.6 & \underline{84.6} & 17.6 & 46.2 & \underline{6.7} & 48.9 \\
    %
    & CBST~\cite{zou2018domain} & ResNet-38~\cite{he2016deep} & 87.1 & 43.9 & 89.7 & 14.8 & 47.7 & 85.4 & 90.3 & 45.4 & 26.6 & 85.4 & 20.5 & 49.8 & 10.3 & 53.6 \\
    & AdaptSegNet~\cite{tsai2018learning} & ResNet-101~\cite{he2016deep} & 83.9 & 34.2 & 88.3 & 18.8 & 40.2 & 86.2 & 93.1 & 47.8 & 21.7 & 80.9 & 47.8 & 48.3 & 8.6 & 53.8 \\
    & AdaptSegNet~\cite{tsai2018learning} & ResNet-50~\cite{he2016deep} & 85.4 & 34.6 & 88.1 & 18.9 & 39.1 & 82.3 & 89.1 & 43.2 & 22.4 & 79.9 & 44.6 & 46.0 & 5.3 & 52.2 \\
    & Ours w/o $\mathcal{L}_\mathrm{consis}$ & ResNet-50~\cite{he2016deep} & 84.4 & 31.2 & 87.7 & 18.6 & 38.0 & 80.7 & 85.4 & 43.5 & 19.8 & 79.4 & {45.3} & 44.2 & 5.1 & 51.0 \\
    & Ours & ResNet-50~\cite{he2016deep} & 90.2 & 37.2 & 91.2 & 22.0 & 41.1 & 86.3 & 91.7 & 47.1 & 25.1 & 83.0 & 48.0 & 47.5 & 6.2 & 55.1 \\
    \midrule
    \multirow{6}{*}{Rio} & Cross-City~\cite{chen2017no} & - & 74.2 & 43.9 & 79.0 & 2.4 & 7.5 & 77.8 & 69.5 & 39.3 & 10.3 & 67.9 & 41.2 & 27.9 & 10.9 & 42.5 \\
    %
    % & Synth.~\cite{zou2018domain} & ResNet-38~\cite{he2016deep} & \underline{80.6} & 36.0 & 81.8 & \textbf{21.0} & 33.1 & 79.0 & 64.7 & 36.0 & 21.0 & 73.1 & \underline{33.6} & 22.5 & 7.8 & 45.4 \\
    %
    % & ST~\cite{zou2018domain} & ResNet-38~\cite{he2016deep} & 80.1 & 41.4 & \underline{83.8} & 19.1 & \textbf{39.1} & \textbf{80.8} & 71.2 & \underline{56.3} & 27.7 & \textbf{79.9} & 32.7 & 36.4 & 12.2 & 50.8 \\
    % %
    & CBST~\cite{zou2018domain} & ResNet-38~\cite{he2016deep} & 84.3 & 55.2 & 85.4 & 19.6 & 30.1 & 80.5 & 77.9 & 55.2 & 28.6 & 79.7 & 33.2 & 37.6 & 11.5 & 52.2 \\
    & AdaptSegNet~\cite{tsai2018learning} & ResNet-101~\cite{he2016deep} & 76.2 & 44.7 & 84.6 & 9.3 & 25.5 & 81.8 & 87.3 & 55.3 & 32.7 & 74.3 & 28.9 & 43.0 & 27.6 & 51.6 \\
    & AdaptSegNet~\cite{tsai2018learning} & ResNet-50~\cite{he2016deep} & 75.8 & 43.9 & 80.7 & 7.7 & 21.1 & 80.8 & 88.0 & 51.2 & 27.4 & 71.1 & 25.6 & 43.7 & 26.9 & 49.5 \\
    & Ours w/o $\mathcal{L}_\mathrm{consis}$ & ResNet-50~\cite{he2016deep} & 74.7 & 44.1 & 81.2 & 5.3 & 19.2 & 80.7 & 86.3 & 52.3 & 27.7 & 69.2 & 24.1 & 45.4 & 25.2 & 48.9 \\
    & Ours & ResNet-50~\cite{he2016deep} & 77.5 & 43.3 & 81.2 & 10.1 & 23.2 & 79.7 & 88.2 & 57.4 & 31.9 & 72.2 & 29.1 & 38.9 & 22.4 & 50.4 \\
    \midrule
    \multirow{6}{*}{Tokyo} & Cross-City~\cite{chen2017no} & - & 83.4 & 35.4 & 72.8 & 12.3 & 12.7 & 77.4 & 64.3 & 42.7 & 21.5 & 64.1 & 20.8 & 8.9 & 40.3 & 42.8 \\
    %
    % & Synth.~\cite{zou2018domain} & ResNet-38~\cite{he2016deep} & \underline{83.8} & 26.4 & 73.0 & 6.5 & 27.0 & 80.5 & 46.6 & 35.6 & 22.8 & 71.3 & 4.2 & 10.5 & 36.1 & 40.3 \\
    %
    % & ST~\cite{zou2018domain} & ResNet-38~\cite{he2016deep} & 83.1 & 27.7 & 74.8 & 7.1 & \underline{29.4} & \textbf{84.4} & 48.5 & \textbf{57.2} & 23.3 & \underline{73.3} & 3.3 & 22.7 & 45.8 & 44.6 \\
    % %
    & CBST~\cite{zou2018domain} & ResNet-38~\cite{he2016deep} & 85.2 & 33.6 & 80.4 & 8.3 & 31.1 & 83.9 & 78.2 & 53.2 & 28.9 & 72.7 & 4.4 & 27.0 & 47.0 & 48.8 \\
    & AdaptSegNet~\cite{tsai2018learning} & ResNet-101~\cite{he2016deep} & 81.5 & 26.0 & 77.8 & 17.8 & 26.8 & 82.7 & 90.9 & 55.8 & 38.0 & 72.1 & 4.2 & 24.5 & 50.8 & 49.9 \\
    & AdaptSegNet~\cite{tsai2018learning} & ResNet-50~\cite{he2016deep} & 76.0 & 25.3 & 78.1 & 15.4 & 22.3 & 81.3 & 91.1 & 45.2 & 34.6 & 69.3 & 2.3 & 20.7 & 48.2 & 46.9 \\
    & Ours w/o $\mathcal{L}_\mathrm{consis}$ & ResNet-50~\cite{he2016deep} & 72.3 & 24.9 & 77.6 & 14.3 & 23.1 & 80.9 & 90.7 & 43.6 & 35.2 & 68.9 & 3.1 & 19.8 & 42.4 & 45.9 \\
    & Ours & ResNet-50~\cite{he2016deep} & 82.1 & 29.3 & 78.2 & 18.2 & 27.5 & 83.1 & 91.2 & 56.4 & 37.8 & 74.3 & 9.5 & 26.0 & 52.1 & 51.2 \\
    \midrule
    \multirow{6}{*}{Taipei} & Cross-City~\cite{chen2017no} & - & 78.6 & 28.6 & 80.0 & 13.1 & 7.6 & 68.2 & 82.1 & 16.8 & 9.4 & 60.4 & 34.0 & 26.5 & 9.9 & 39.6 \\
    %
    % & Synth.~\cite{zou2018domain} & ResNet-38~\cite{he2016deep} & \underline{84.9} & 26.0 & 80.1 & 8.3 & \textbf{28.0} & 73.9 & 54.4 & 18.9 & 26.8 & 71.6 & 26.0 & 48.2 & 14.7 & 43.2 \\
    % %
    % & ST~\cite{zou2018domain} & ResNet-38~\cite{he2016deep} & 83.1 & 23.5 & 78.2 & 9.6 & \underline{25.4} & \underline{74.8} & 35.9 & \textbf{33.2} & \underline{27.3} & \underline{75.2} & 32.3 & \underline{52.2} & \underline{28.8} & 44.6 \\
    %
    & CBST~\cite{zou2018domain} & ResNet-38~\cite{he2016deep} & 86.1 & 35.2 & 84.2 & 15.0 & 22.2 & 75.6 & 74.9 & 22.7 & 33.1 & 78.0 & 37.6 & 58.0 & 30.9 & 50.3 \\
    & AdaptSegNet~\cite{tsai2018learning} & ResNet-101~\cite{he2016deep} & 81.7 & 29.5 & 85.2 & 26.4 & 15.6 & 76.7 & 91.7 & 31.0 & 12.5 & 71.5 & 41.1 & 47.3 & 27.7 & 49.1 \\
    & AdaptSegNet~\cite{tsai2018learning} & ResNet-50~\cite{he2016deep} & 81.8 & 27.8 & 83.2 & 24.4 & 12.6 & 74.1 & 88.7 & 30.9 & 11.1 & 70.8 & 40.2 & 45.3 & 26.2 & 47.5 \\
    & Ours w/o $\mathcal{L}_\mathrm{consis}$ & ResNet-50~\cite{he2016deep} & 79.6 & 26.9 & 84.1 & 23.7 & 14.1 & 72.8 & 86.5 & 30.3 & 9.9 & 69.9 & 40.6 & 44.7 & 25.8 & 46.8 \\
    & Ours & ResNet-50~\cite{he2016deep} & 79.7 & 28.1 & 85.1 & 24.4 & 16.4 & 74.3 & 87.9 & 29.5 & 12.8 & 69.8 & 40.0 & 46.8 & 28.1 & 47.9 \\
    \bottomrule
    \end{tabular}
    }
    \end{center}
    \vspace{3.0\tablemargin}
  \end{table*}

\subsubsection{SYNTHIA to Cityscapes}

% \vspace{\paramargin}
\paragraph{Dataset.} 
%
% To adapt from the SYNTHIA~\cite{ros2016synthia} dataset to the Cityscapes~\cite{cordts2016cityscapes} dataset, 
We use the SYNTHIA-RAND-CITYSCAPES~\cite{ros2016synthia} set as the source domain which contains $9,400$ images compatible with the Cityscapes annotated classes. 
Following Dundar~\etal~\cite{dundar2018domain}, we evaluate images on the Cityscapes validation set with $16$ classes.
%
% \revised{
%  we evaluate our proposed method using the DRN-$26$~\cite{yu2017dilated} and the FCN$8$s-VGG$16$~\cite{long2015fully}.
% }

\vspace{\paramargin}
\paragraph{Results.}
We compare our approach with the state-of-the-art methods~\cite{tobin2017domain,zhu2017unpaired,liu2017unsupervised,hoffman2016fcns,dundar2018domain}.
The bottom block of Table~\ref{table:exp-seg-syn2real} presents the experimental results.
In either DRN-$26$~\cite{yu2017dilated} or FCN$8$s~\cite{long2015fully} backbone, our method achieves state-of-the-art performance.
% \revised{
% When using the DRN-$26$~\cite{yu2017dilated}, our method compares favorably against the state-of-the-art methods, outperforming the previous best competitor~\cite{Huang_2018_ECCV} by $3.9$\% in mean IoU and $3.0$\% in pixel accuracy.
% %
% A similar observation can be observed when employing the FCN$8$s-VGG$16$~\cite{long2015fully} as the task network.
% }
%
Likewise, we show sizable improvement using the proposed cross-domain consistency loss $\mathcal{L}_\mathrm{consis}$.

% with the proposed cross-domain consistency loss $\mathcal{L}_\mathrm{consis}$, our method effectively improves the mean IoU by $3.6$\% and the pixel accuracy by $4.2$\%.

\begin{table*}[t]
  \scriptsize
  \ra{1.2}
  \begin{center}
    \caption{\textbf{Synthetic-to-real (\texttt{SUNCG} $\rightarrow$ \texttt{NYUv2}) adaptation for depth prediction.}
    %
    % Adaptation: \texttt{SUNCG} $\rightarrow$ \texttt{NYUv2}. 
    %
    The column ``Supervision'' indicates methods trained with NYUv2 training data.
    % that method is learned in a supervised fashion.
    %\jb{What real image pairs? Yun-Chun: NYUDv2. Namely, these are supervised learning methods.}
    %
    We denote the top two results as \tb{bold} and \underline{underlined}.
    % numbers indicate the top two results, respectively.
%\jb{
%1) Group methods into different blocks, e.g., methods trained with the NYUv2 training set and methods that do not do that. 
%2) Move the baseline to the top. 
%3) Move Eigen's methods together. What's "fine"?
%4) Show ours with/without results. Reviewer would like to know how much did our contribution help. Not just show the final numbers.
%5) NYUv2 testing set? Which spilt?}
    }
    \vspace{\tablecapmargin}
    \label{table:exp-depth}
    \resizebox{\textwidth}{!} 
    {
    \begin{tabular}{l|c|cccc|ccc}
    \toprule
    Method & Supervision & Abs. Rel. $\downarrow$ & Sq. Rel. $\downarrow$ & RMSE $\downarrow$ & RMSE log. $\downarrow$ & $\delta < 1.25$ $\uparrow$ & $\delta < 1.25^2$ $\uparrow$ & $\delta < 1.25^3$ $\uparrow$ \\
    \midrule
    %
    % Ladicky~\etal~\cite{ladicky2014pulling} & $\checkmark$ & - & - & - & - & 0.542 & 0.829 & 0.940 \\
    %
    Liu~\etal~\cite{liu2016learning} & $\checkmark$ & 0.213 & - & 0.759 & - & 0.650 & 0.906 & 0.976 \\
    Eigen~\etal~\cite{eigen2014depth} Fine & $\checkmark$ & 0.215 & 0.212 & 0.907 & 0.285 & 0.611 & 0.887 & 0.971 \\
    Eigen~\etal~\cite{eigen2015predicting} (VGG) & $\checkmark$ & 0.158 & 0.121 & 0.641 & 0.214 & 0.769 & 0.950 & 0.988 \\
    T$^2$Net~\cite{zheng2018t2net} & $\checkmark$ & 0.157 & 0.125 & 0.556 & 0.199 & 0.779 & 0.943 & 0.983 \\
    \midrule
    Synth. & & 0.304 & 0.394 & 1.024 & 0.369 & 0.458 & 0.771 & 0.916 \\
    Baseline (train set mean) & & 0.439 & 0.641 & 1.148 & 0.415 & 0.412 & 0.692 & 0.586 \\
    T$^2$Net~\cite{zheng2018t2net} & & 0.257 & \underline{0.281} & 0.915 & \underline{0.305} & 0.540 & 0.832 & \underline{0.948} \\
    %
    % \midrule
    %
    Ours w/o $\mathcal{L}_\mathrm{consis}$  & & \underline{0.254} & 0.283 & \underline{0.911} & 0.306 & \underline{0.541} & \underline{0.835} & 0.947 \\
    Ours     & & \textbf{0.233} & \textbf{0.272} & \textbf{0.898} & \textbf{0.289} & \textbf{0.562} & \textbf{0.853} & \textbf{0.952} \\
    \bottomrule
    \end{tabular}
    }
    \end{center}
    \vspace{3.0\tablemargin}
  \end{table*}

\subsubsection{Cityscapes to Cross-City}

% \vspace{\paramargin}
\paragraph{Dataset.} 
In addition to the \emph{synthetic-to-real} adaptation, we conduct an experiment on the Cross-City dataset~\cite{chen2017no} which is a \emph{real-to-real} adaptation.
The dataset contains four different cities: Rio, Rome, Tokyo, and Taipei, where each city has $3,200$ images without annotations and $100$ images with pixel-level ground truths for $13$ classes.
Following Tsai~\etal~\cite{tsai2018learning}, we use the Cityscapes~\cite{cordts2016cityscapes} training set as our source domain and adapt the model to each target city using $3,200$ images, and use the $100$ annotated images for evaluation.

\vspace{\paramargin}
\paragraph{Results.}
%
% \revised{
We compare our approach with the Cross-City~\cite{chen2017no}, the CBST~\cite{zou2018domain}, and the AdaptSegNet~\cite{tsai2018learning}. 
% }
%
Table~\ref{table:exp-seg-real2real} shows that our method achieve state-of-the-art performance on two out of four cities.
%
% Due to the limited GPU memory, we use the ResNet-$50$~\cite{he2016deep} as our feature backbone.
%
% We observe that our method achieve state-of-the-art performance on two out of four cities.
%
Note that the results in AdaptSegNet~\cite{tsai2018learning} are obtained by using a ResNet-$101$~\cite{he2016deep}.
We run their publicly available code with the default settings and report the results using the ResNet-$50$~\cite{he2016deep} as the feature backbone for a fair comparison.
Under the same experimental setting, our approach compares favorably against state-of-the-art methods.
Furthermore, we show that enforcing cross-domain consistency constraints, our method effectively and consistently improves the results evaluated on all four cities.

\subsection{Single-view depth estimation}

To show that our formulation is not limited to semantic segmentation, we present experimental results for single-view depth prediction task.
Specifically, we use SUNCG~\cite{song2017semantic} as the source domain and adapt the model to the NYUDv2~\cite{silberman2012indoor} dataset.
%
% We show evaluations of the model trained on a synthetic dataset (\ie SUNCG~\cite{song2017semantic}) and test the adapted model on real-world images from the NYUDv2~\cite{silberman2012indoor} dataset.

\vspace{\paramargin}
\paragraph{Dataset.} 
To generate the paired synthetic training data, we rendered RGB images and depth map from the SUNCG dataset~\cite{song2017semantic}, which contains $45,622$ 3D houses with various room types. 
Following Zheng~\etal~\cite{zheng2018t2net}, we choose the camera locations, poses and parameters based on the distribution of real NYUDv2 dataset~\cite{silberman2012indoor} and retain valid depth maps using the criteria described by Song~\etal~\cite{song2017semantic}.
In total, we generate $130,190$ valid views from $4,562$ different houses.

\vspace{\paramargin}
\paragraph{Evaluation protocols.}

We use the root mean square error (RMSE) and the log scale version (RMSE log.), the squared relative difference (Sq. Rel.) and the absolute relative difference (Abs. Rel.), and the accuracy measured by thresholding ($\delta$ $<$ threshold).

\vspace{\paramargin}
\paragraph{Task network.} 
We initialize our task network from the unsupervised version of Zheng~\etal~\cite{zheng2018t2net}.

\vspace{\paramargin}
\paragraph{Results.}
Table~\ref{table:exp-depth} shows the comparisons with prior methods~\cite{liu2016learning,eigen2014depth,eigen2015predicting,zheng2018t2net}.
Here, the column ``Supervision'' indicates that the method is learned in a supervised fashion.
While not directly comparable, we report their results for completeness.
Under the same experimental settings, we observe that our method achieves state-of-the-art performance on all adopted evaluation metrics.
Moreover, with the integration of the cross-domain consistency loss $\mathcal{L}_\mathrm{consis}$, our method shows consistently improved performance.

\subsection{Optical flow estimation}

% To further show the applicability of the proposed method, we conduct experiments on optical flow estimation.
%
We show evaluations of the model trained on a synthetic dataset (\ie MPI Sintel~\cite{butler2012naturalistic}) and test the adapted model on real-world images from the KITTI 2012~\cite{geiger2012we} and KITTI 2015~\cite{menze2015object} datasets.

\vspace{\paramargin}
\paragraph{Dataset.}
The MPI Sintel dataset~\cite{butler2012naturalistic} consists of $1,401$ images rendered from artificial scenes.
% with special attention to realistic image properties.
%
There are two versions: 
1) the final version consists of images with motion blur and atmospheric effects, and
2) the clean version does not include these effects.
We use the clean version as the source dataset.
We report two results obtained by 1) using the KITTI 2012~\cite{geiger2012we} as the target dataset and 2) using the KITTI 2015~\cite{menze2015object} as the target dataset.

\vspace{\paramargin}
\paragraph{Evaluation protocols.}
We adopt the average endpoint error (AEPE) and the F1 score for both KITTI 2012 and KITTI 2015 to evaluate the performance.

\vspace{\paramargin}
\paragraph{Task network.}
Our task network is initialized from the PWC-Net~\cite{sun2018pwc} (without finetuning on the KITTI dataset).

\vspace{\paramargin}
\paragraph{Results.}
We compare our approach with the state-of-the-art methods~\cite{sun2018pwc,ranjan2017optical,ilg2017flownet}.
Table~\ref{table:exp-flow} shows that our method achieves improved performance on both datasets.
% achieves the state-of-the-art performance in terms of AEPE on both dataset.
%
% We note that the column ``finetune'' indicates that the method is finetuned on the KITTI dataset.
%
% While not directly comparable, we report their results for completeness.
%
% Under the same experimental settings, our method achieves the state-of-the-art performance in terms of AEPE on both dataset.
%
When incorporating the proposed cross-domain consistency loss $\mathcal{L}_\mathrm{consis}$, our model improves the results by $1.76$ in terms of average endpoint error on the KITTI 2012 test set and $10.6\%$ in terms of F1-all on the KITTI 2015 test set.

\begin{table}[t]
  \scriptsize
  \ra{1.2}
  \begin{center}
    \caption{\textbf{Experimental results of synthetic-to-real adaptation for optical flow estimation.}
    Left: \texttt{MPI Sintel} $\rightarrow$ \texttt{KITTI 2012}.
    Right: \texttt{MPI Sintel} $\rightarrow$ \texttt{KITTI 2015}.
    The column ``finetune'' indicates that method is finetuned on the KITTI dataset.
    The bold and the underlined numbers indicate top two results, respectively.}
    \label{table:exp-flow}
    \vspace{\tablecapmargin}
    \resizebox{\linewidth}{!} 
    {
    \begin{tabular}{l|c|ccc|ccc}
    \toprule
    \multirow{3}{*}{Method} & \multirow{3}{*}{finetune} & \multicolumn{3}{c|}{KITTI 2012} & \multicolumn{3}{c}{KITTI 2015} \\
    & & AEPE & AEPE & F$1$-Noc & AEPE & F$1$-all & F$1$-all \\
    & & \emph{train} & \emph{test} & \emph{test} & \emph{train} & \emph{train} & \emph{test} \\
    \midrule
    SpyNet~\cite{ranjan2017optical} & $\checkmark$ & 4.13 & 4.7 & 12.31\% & - & - & 35.05\% \\
    FlowNet2~\cite{ilg2017flownet} & $\checkmark$ & 1.28 & 1.8 & 4.82\% & 2.30 & 8.61\% & 10.41\% \\
    PWC-Net~\cite{sun2018pwc} & $\checkmark$ & 1.45 & 1.7 & 4.22\% & 2.16 & 9.80\% & 9.60\% \\
    \midrule
    FlowNet2~\cite{ilg2017flownet} & & 4.09 & - & - & \underline{10.06} & \underline{30.37}\% & - \\
    PWC-Net~\cite{sun2018pwc} & & \underline{4.14} & \underline{4.22} & \textbf{8.10\%} & 10.35 & 33.67\% & - \\
    Ours w/o $\mathcal{L}_\mathrm{consis}$ & & 4.16 & 4.92 & 13.52\% & 10.76 & 34.01\% & \underline{36.43}\% \\
    Ours     & & \textbf{2.19} & \textbf{3.16} & \underline{8.57}\% & \textbf{8.02} & \textbf{23.14}\% & \textbf{25.83}\% \\
    \bottomrule
    \end{tabular}
    }
  \end{center}
  \vspace{3.5\tablemargin}
\end{table}

\subsection{Limitations}

Our method is memory-intensive as the training involves multiple networks at the same time.
Potential approaches to alleviate this issue include 1) adopting partial sharing on the two task networks, \eg share the last few layers of the two task networks, and 2) sharing the encoders in the image translation network (\ie $G_{S \rightarrow T}$ and $G_{T \rightarrow S}$).

\section{Conclusions} \label{sec:conclusions}

We have presented a simple yet surprisingly effective loss for improving pixel-level unsupervised domain adaption for dense prediction tasks. 
We show that by incorporating the proposed cross-domain consistency loss, our method consistently improves the performances over a wide range of tasks.
Through extensive experiments, we demonstrate that our method is applicable to a wide variety of tasks. 

\vspace{\paramargin}
\paragraph{Acknowledgement.}
This work was supported in part by NSF under Grant No. 1755785,  No. 1149783, Ministry of Science and Technology (MOST) under grants 107-2628-E-001-005-MY3 and 108-2634-F-007-009, and gifts from Adobe, Verisk, and NEC. We thank the support of NVIDIA Corporation with the GPU donation.

{\small
\bibliographystyle{ieee}
\bibliography{reference}
}

\end{document}